\documentclass[11pt,twoside]{article}
\pdfoutput=1 
\usepackage{fullpage}

\usepackage{amsmath}
\usepackage{bm}
\usepackage{bbm}

\usepackage{algorithm}
\usepackage{algorithmic}

\usepackage{hyperref}       
\usepackage{url}            
\usepackage{booktabs}       
\usepackage{amsfonts,amssymb, amsthm}       
\usepackage{nicefrac}       
\usepackage{microtype}      
\usepackage{xcolor}         

\usepackage{graphicx}
\usepackage{caption}
\usepackage{subcaption}
\usepackage{float}
\usepackage{bm}
\usepackage{tabularx}
\usepackage{wrapfig}
\usepackage{comment}
\usepackage{cancel}
\usepackage{enumitem}
\usepackage{xcolor}
\usepackage[normalem]{ulem}
\usepackage{xspace, cleveref, dsfont}
\usepackage{breakcites}
\usepackage{svg}
\usepackage[numbers]{natbib}
\usepackage{mathtools}

\newcommand{\va}{\bm{a}}

\newcommand{\vv}{\bm{v}}
\newcommand{\vx}{\bm{x}}

\newcommand{\vzero}{\bm{0}}
\newcommand{\vone}{\bm{1}}

\newcommand{\mA}{\bm{A}}
\newcommand{\mB}{\bm{B}}

\newcommand{\mId}{\bm{I}}

\newcommand{\mQ}{\bm{Q}}

\newcommand{\mU}{\bm{U}}
\newcommand{\mL}{\bm{L}}

\newcommand{\mX}{\bm{X}}
\newcommand{\mDelta}{\bm{\Delta}}

\newcommand{\mSigma}{\bm{\Sigma}}

\newcommand{\calA}{\mathcal{A}}

\newcommand{\calE}{\mathcal{E}}

\newcommand{\calG}{\mathcal{G}}

\newcommand{\calL}{\mathcal{L}}

\newcommand{\calN}{\mathcal{N}}

\newcommand{\calS}{\mathcal{S}}

\newcommand{\bbR}{\mathbb{R}}
\newcommand{\bbS}{\mathbb{S}}

\DeclareMathOperator*{\argmin}{arg\,min}

\DeclarePairedDelimiter{\norm}{\lVert}{\rVert}
\newcommand{\fronorm}[1]{\lVert #1 \rVert_F}
\newcommand{\nucnorm}[1]{\lVert #1 \rVert_{*}}
\newcommand{\opnorm}[1]{\left\lVert #1 \right\rVert_{\mathrm{op}}}
\newcommand{\normtwo}[1]{\norm*{#1}_2}
\let\normfro\fronorm
\let\normop\opnorm
\let\normnuc\nucnorm

\newcommand*\dd{\mathop{}\!\mathrm{d}}

\DeclarePairedDelimiterX{\ip}[2]{\langle}{\rangle}{#1,#2}
\DeclarePairedDelimiter{\abs}{\lvert}{\rvert}

\renewcommand{\P}[1]{\mathbb{P}\left(#1\right)}
\newcommand{\E}[1]{\mathbb{E}\left[#1\right]}
\newcommand{\var}{\operatorname{var}}
\newcommand{\indicator}{\mathds{1}}

\newtheorem{theorem}{Theorem}

\newtheorem{assumption}{Assumption}
\newtheorem{proposition}{Proposition}
\newtheorem{lemma}{Lemma}
\newtheorem{corollary}{Corollary}

\newtheorem{remark}{Remark}

\newcommand{\queryabbr}{\text{PAQ}\xspace}
\newcommand{\queryabbrs}{\text{PAQs}\xspace}
\newcommand{\queryname}{\text{perceptual adjustment query}\xspace}

\newcommand{\idxitem}{i}
\newcommand{\numitems}{n}

\newcommand{\boundary}{y}
\newcommand{\boundarystar}{\boundary_{\star}}
\newcommand{\boundaryup}{\boundary^\uparrow}
\newcommand{\boundaryupstar}{\boundaryup_{\star}}
\newcommand{\trunc}{\tau}
\let\thresh\trunc

\newcommand{\scale}{\gamma}
\newcommand{\scalebar}{\bar{\scale}}
\newcommand{\scaletilde}{\widetilde{\scale}}
\newcommand{\scalestar}{\scale_\star}

\newcommand{\noise}{\eta}
\newcommand{\noisebar}{\bar{\noise}}

\newcommand{\noiseup}{\noise^\uparrow}
\newcommand{\noisevar}{\nu_{\noise}^2}

\newcommand{\noisestar}{\noise_{\star}}
\newcommand{\noiseupstar}{\noiseup_{\star}}
\newcommand{\noisevarstar}{\nu_{\noise,\star}^2}
\newcommand{\noisemedian}{\mu_\boundary}
\newcommand{\noisemedianstar}{\noisemedian^\star}

\newcommand{\Sigstar}{\mSigma^\star}
\newcommand{\Sighat}{\widehat{\mSigma}}

\newcommand{\normsig}[1]{\norm{#1}_{\mSigma}}
\newcommand{\normsigstar}[1]{\norm{#1}_{\Sigstar}}
\newcommand{\fourthmoment}{M}

\newcommand{\mAtilde}{\widetilde{\mA}}
\newcommand{\mAbar}{\bar{\mA}}
\newcommand{\mAinv}{\mA^{\sf inv}}

\newcommand{\constscale}{c_{\sf scale}}

\newcommand{\opa}[1]{\va_{#1}\va_{#1}^\top} 
\newcommand{\quadSig}[1]{\va_{#1}^\top\Sigstar\va_{#1}}
\newcommand{\quadmU}[1]{\va_{#1}^\top\mU\va_{#1}}

\newcommand{\truncprime}{\trunc^\prime}

\newcommand{\mAtrunc}{\mAtilde^{\truncprime}}
\DeclareMathOperator{\trace}{tr}
\newcommand{\tr}[1]{\trace\left(#1\right)}

\newcommand{\stepone}{\text{(i)}\xspace}
\newcommand{\steptwo}{\text{(ii)}\xspace}
\newcommand{\stepthree}{\text{(iii)}\xspace}
\newcommand{\stepfour}{\text{(iv)}\xspace}

\newcommand{\defn}{:=}
\newcommand{\const}{c}
\newcommand{\Const}{C}

\newcommand{\varnoise}{\noisevar}

\newcommand{\sphere}{\calS}

\let\normal\calN
\let\reals\bbR
\newcommand{\Prob}{\mathbb{P}}

\newcommand{\sample}{\sim}

\newcommand{\lessorder}{\lesssim}
\newcommand{\gtrorder}{\gtrsim}
\newcommand{\idmtx}{\mId}
\newcommand{\dimension}{d}
\newcommand{\Expect}{\mathbb{E}}
\newcommand{\sigularval}{\sigma}
\newcommand{\rank}{r}
\newcommand{\singularvalmin}{\sigularval_\rank}

\newcommand{\setcover}{\calA}

\newcommand{\varmtx}{\mU}

\newcommand{\constbernone}{u_1}
\newcommand{\constberntwo}{u_2}
\newcommand{\power}{p}
\newcommand{\vargauss}{G}

\begin{document}

\begin{center}

  {\bf{\LARGE{Perceptual adjustment queries and an inverted measurement paradigm for low-rank metric learning}}}

\vspace*{.2in}

{\large{
\begin{tabular}{ccccc}
 & & Austin Xu$^\dagger$\qquad  Andrew D. McRae$^\ddagger$\qquad Jingyan Wang$^\star$ & &\\
 & & Mark Davenport$^\dagger$\qquad Ashwin Pananjady$^{\star, \dagger}$ & &
\end{tabular}
}}
\vspace*{.2in}

\begin{tabular}{c}
$^\star$School of Industrial and Systems Engineering \\
$^\dagger$School of Electrical and Computer Engineering \\
Georgia Institute of Technology\\
\\
$^\ddagger$ Institute of Mathematics\\ 
Ecole Polytechnique F\'{e}d\'{e}rale de Lausanne (EPFL)
\end{tabular}

\vspace*{.2in}

September 8, 2023

\vspace*{.2in}

\begin{abstract}
We introduce a new type of query mechanism for collecting human feedback, called the \queryname (\queryabbr). Being both informative and cognitively lightweight, the \queryabbr adopts an inverted measurement scheme, and combines advantages from both cardinal and ordinal queries. We showcase the \queryabbr in the metric learning problem, where we collect \queryabbr measurements to learn an unknown Mahalanobis distance. This gives rise to a high-dimensional, low-rank matrix estimation problem to which standard matrix estimators cannot be applied. Consequently, we develop a two-stage estimator for metric learning from \queryabbrs, and provide sample complexity guarantees for this estimator. We present  numerical simulations demonstrating the performance of the estimator and its notable properties.
\end{abstract}
\end{center}

\section{Introduction}\label{sec:intro}
Should we query cardinal or ordinal data from people? This question arises in a broad range of applications, such as in conducting surveys~\cite{rankin1980system,harzing2009national,yannakakis2011reporting}, grading assignments~\cite{shah2013case,raman2014grading}, evaluating employees~\cite{goffin2011relative}, and comparing or rating products~\cite{barnett2003modern,batley2008ordinal}, to name a few. \emph{Cardinal} data are numerical scores. For example, teachers score writing assignments in the range of $0$-$100$, and survey respondents express their agreement with a statement on a scale of $1$ to $7$. \emph{Ordinal} data are relations between items, such as pairwise comparisons (choosing the better item in a pair) and rankings (ordering all or a subset of items). There is no free lunch, and both cardinal and ordinal queries have pros and cons. 

On the one hand, collecting ordinal data is typically more efficient in terms of worker time and cognitive load~\cite{shah2016topology}, and surprisingly often matches or exceeds the accuracy of cardinal data~\cite{rankin1980system,shah2016topology}. The information contained in ordinal queries, however, is fundamentally limited and lacks expressiveness. For example, pairwise comparisons elicit binary responses where two items are compared against each other, but the  absolute placement of these items with respect to the entire pool is lost. On the other hand, cardinal data are more expressive~\cite{wang2019miscalibration}. For example, assigning two items scores of $1$ and $2$ conveys a very different message from assigning them scores of $9$ and $10$, or $1$ and $10$, although all yield the same pairwise comparison outcome. However, the expressiveness of cardinal data often comes at the cost of miscalibration: Prior work has shown that different people have different scales~\cite{griffin2008calibration}, and even a single person's scale can drift over time (e.g.,~\cite{harik2009generalizability,myford2009drift}). These inter-person and intra-person discrepancies make it challenging to interpret and aggregate raw scores effectively.

The goal of this paper is to study whether one can combine the advantages of cardinal and ordinal queries to achieve the best of both worlds. Specifically, we pose the research question:
\begin{quote}
    \emph{Can we develop a new paradigm for human data elicitation that is expressive, accurate, and cognitively lightweight?}
\end{quote}
Towards this goal, we extract key features of both cardinal and ordinal queries, and propose a new type of query scheme that we term the \emph{\queryname} (\queryabbr). As a thought experiment, consider the task of learning an individual's preferences between modes of transport.
The query can take the following forms:
\begin{itemize}[leftmargin=*]
    \item \textbf{Ordinal:} Do you prefer a \$2 bus ride that takes 40 minutes or a \$25 taxi that takes 10 minutes? 

    \item \textbf{Cardinal:} On a scale of $0$ to $1$, how much do you value a \$2 bus ride that takes 40 minutes? 

    \item \textbf{Proposed approach:} To reach the same level of preference for a \$2 bus trip that takes 40 minutes, a taxi that takes 10 minutes would cost \$$x$.
\end{itemize}

\begin{figure}[ht!]
\centering
\includegraphics[width=0.9\linewidth]{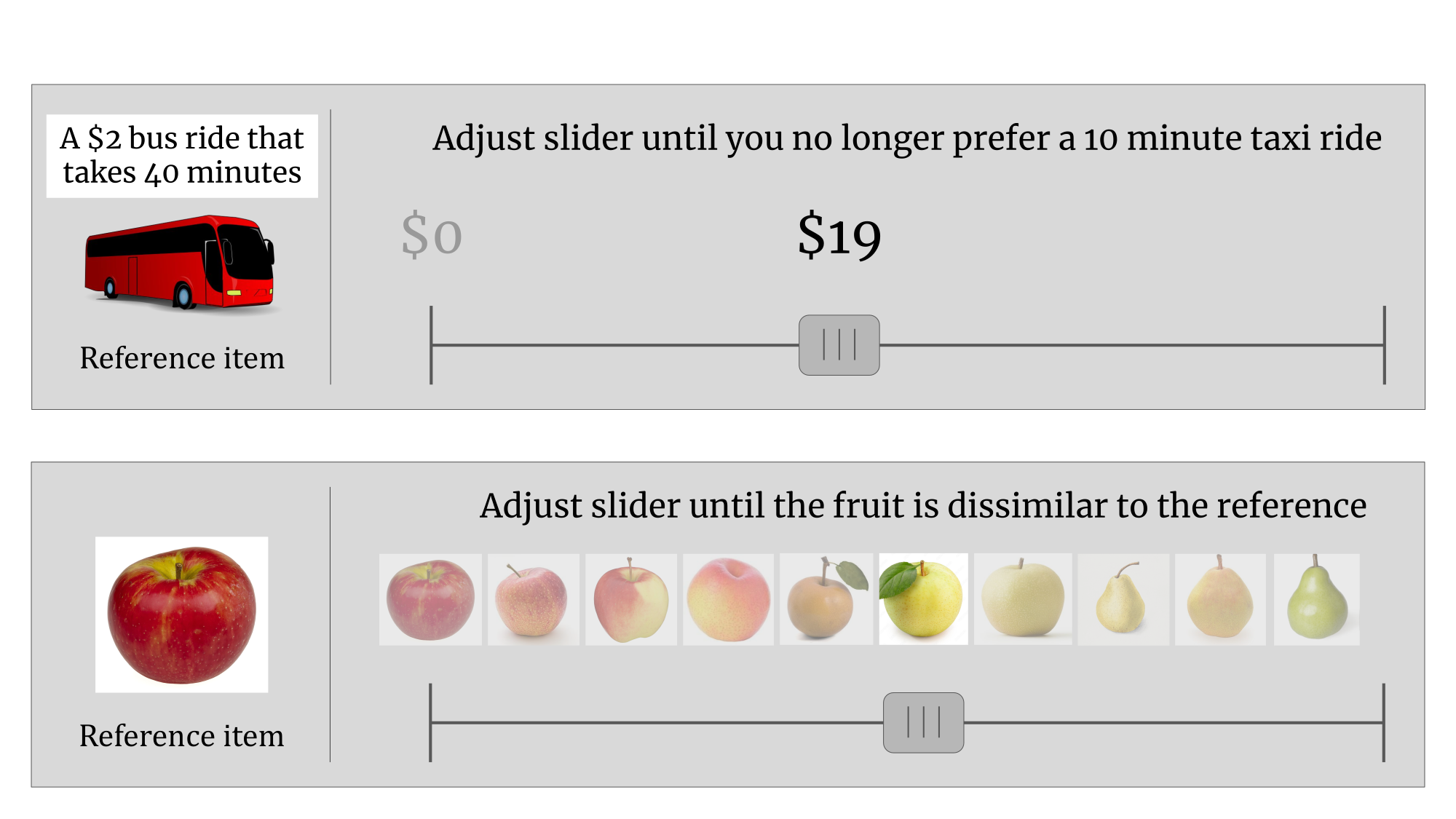}
\caption{The user interface for \queryname (\queryabbr) for preference learning (top) and similarity learning (bottom). \label{fig:ui}}
\end{figure}

A user interface for the proposed approach is shown in Figure~\ref{fig:ui} (top).
We present the user a reference item (a \$2 bus ride that takes 40 minutes), and a sliding bar representing the number of dollars ($x$) for the 10-minute taxi cost. As the user adjusts the slider, the value of $x$ starts with $0$ and gradually increases on a continuous scale. The user is instructed to place the slider at a point where they equally prefer a \$2 bus ride and a taxi ride of $x$ dollars.\footnote{
    The ordinal component is crucial in our proposed \queryname --- we provide a reference item and instruct people to make a relational judgment of the target item compared to the reference item. Hence, the \queryname is distinct from sliding survey questions that elicit purely cardinal responses.
}
The \queryabbr thus combines ordinal and cardinal elicitation in an intuitive fashion: We obtain ordinal information by asking the user to make cognitive judgments in a relative sense by comparing items, and cardinal information can be extracted from the location of the slider. The ordinal reasoning endows the query with accuracy and efficiency, while the cardinal output enables a more expressive response. Moreover, this cardinal output mitigates miscalibration, because instead of asking the user to rate on a subjective and ambiguous notion (i.e., preference), we provide the user a reference object  (i.e., the \$2 bus ride) to anchor their rating scale.

This combination of high per-response information and low cognitive burden makes the deployment of \queryabbrs appealing in a variety of problem settings. For example:
\begin{itemize}
    \item Learning human preferences. As illustrated in the taxi and bus example in~\Cref{fig:ui}, one can ask users to pinpoint the cost at which a taxi ride is equally preferred to the bus ride. In a more complex setting, such as housing preferences, moving the slider can change multiple attributes, such as price, square footage, maintenance fees, proximity to employment, etc. User responses to \queryabbrs yield information-dense statements about how features jointly impact human preferences.
    
    \item Learning a model for color perception. Imagine a user with red-green color blindness, the extent of which we wish to learn. We can present the user with an image of a red square and a sequence of colors that slowly transitions from red to green, and ask them to drag the slider until they perceive a difference in colors. In such a setting, \queryabbrs present users with context (the full sequence of colors and the reference color) to help them indicate their color sensitivity: At what point can you start distinguishing the two colors?
    
    \item Studying generative models. Imagine we wish to characterize how the semantic characteristics of synthesized items (e.g., images) change along different directions of a given generative model. By traversing a continuous path in the model's latent space and generating a corresponding item for each point, \queryabbrs present users with a sequence of items. Using an item at the beginning of the sequence as the reference item, we ask users to mark the first item along the sequence that is semantically different in a meaningful way. For example, to characterize how different directions in the latent space impact breed for a model trained to synthesize images of dogs, we ask users to mark the first image where the breed clearly changes.
\end{itemize}

Beyond combining the strengths of cardinal and ordinal queries, \queryabbrs have additional advantages that are well illustrated with the example in Figure~\ref{fig:ui} (bottom). First, \queryabbrs provide users with the \emph{context} of a specific (continuous) dimension along which items vary. For example, consider a pairwise comparison between the reference item and the ``yellow apple '' selected in Figure~\ref{fig:ui}. They have similar shapes, but different colors. If these two items are shown to the user in isolation, the user lacks context to judge whether they should be considered similar or dissimilar. In contrast, the full spectrum provided in \queryabbrs tells the user that the similarity judgment is apples vs.\ pears. 
The access to such context improves self-consistency in user responses~\cite{canal2020tuplewise}. 
Second, \queryabbrs provide ``hard examples'' by design and thus enable effective learning. Consider Figure~\ref{fig:ui} (bottom): Items on the left of the spectrum  are apples (clearly similar to the reference), and items on the right are pears (clearly dissimilar to the reference), and only a small subset of items in the middle appear ambiguous. \queryabbrs collect information precisely about ``confusing" items in this ambiguous region. On the other hand, if ordinal queries are constructed by selecting uniformly at random from the items shown, an item in the ambiguous region will rarely be presented to the user. 

In this paper, we apply the \queryabbr scheme in the framework of metric learning for human perception. In this problem, items are represented by points in a (possibly high-dimensional) space, and the goal is to learn a distance metric such that a smaller distance between a pair of items means that they are semantically and perceptually closer, and vice versa. Figure~\ref{fig:ui} (bottom) presents a \queryabbr for collecting similarity data for metric learning, where the user is instructed to place the slider at the precise point where the object appears to transition from being similar to dissimilar. 

To construct a sequence of images as shown in Figure~\ref{fig:ui} (bottom), one can traverse a path in the latent space of a generative model --- given a latent feature vector, the generative model synthesizes a corresponding image. In other settings, such as the taxi example in Figure~\ref{fig:ui} (top) or the housing preference task mentioned above, a sequence of items can be formed by gradually changing the value of interpretable features, such as price and square footage. 

\subsection{Do \queryabbrs improve upon ordinal queries? A simulation vignette}

\begin{figure}[tb]
    \centering
    \includegraphics[width=0.45\linewidth]{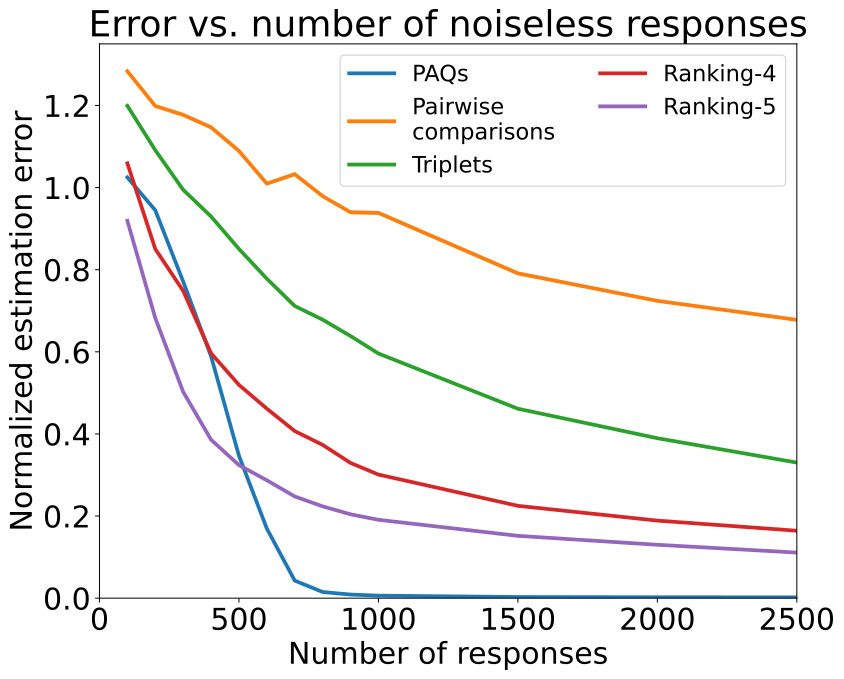}
    \caption{Simulation comparing performance of noiseless responses to \queryabbrs and various ordinal queries when applied to low-rank metric learning. Ranking-$k$ denotes that $k$ items are ranked in terms of similarity to a reference item. For each query type, we plot the mean and standard error of the mean (shaded regions, not visible) of the normalized estimation error $\fronorm{\Sigstar - \Sighat} / \fronorm{\Sigstar}$ over $10$ independent trials.} 
    \label{fig:noiseless_sims}
\end{figure}

Consider the problem of Mahalanobis metric learning, which forms the focus of this paper. In this setting, items are represented as points in the vector space $\bbR^d$, which is in turn endowed with a Mahalanobis metric parametrized by a symmetric positive semidefinite matrix $\Sigstar \in \bbR^{d \times d}$. The (dis-)similarity of two items is determined by their distance under the metric: The larger the (squared) distance $\normsigstar{\vx - \vx^\prime}^2 = (\vx - \vx^\prime)^\top\Sigstar(\vx-\vx^\prime)$ between two items $\vx$ and $\vx^\prime$ is, the more dissimilar the items are. We are particularly interested in the setting in which $\Sigstar$ is \emph{low-rank}, which covers several important settings. For example, a user may make preference judgements using a small number of interpretable features~\cite{xu2020simultaneous, canal2022one}. For another example, it has been shown that a small number of linear directions capture a vast majority of semantic changes in the latent space of a popular generative model, StyleGAN2~\cite{harkonen2020ganspace}.

Established approaches in metric learning use ordinal queries, such as pairwise comparisons (``Are items $\vx$ and $\vx^\prime$ similar?'')~\cite{ying2009sparse, bian2012constrained, guo2014guaranteed, bellet2015robustness}, triplet comparisons~\cite{mason2017learning} (``Which of the two items $\vx_1$ and $\vx_2$ is closer to reference item $\vx_0$?''), and ranking queries (``Given a reference item $\vx_0$, rank the set of items $\vx_1, \ldots, \vx_k$ in terms of similarity to $\vx_0$'')~\cite{canal2020tuplewise}. In Figure~\ref{fig:noiseless_sims}, we simulate the performance of such queries in a toy metric learning setup against the performance of PAQs. 

In particular, we choose a random low-rank matrix $\Sigstar$ in dimension $50$ with rank $10$ (see Appendix~\ref{app:sims} for our precise construction, which resembles the setup of~\cite{chen2015exact}) and use the models of~\cite{mason2017learning,canal2020tuplewise} to produce standard pairwise, triplet, and ranking-$k$ queries. We also use state-of-the-art algorithms to estimate the low-rank metric from these types of queries~\cite{mason2017learning,canal2020tuplewise}. In addition to these ordinal queries, we simulate \queryabbr responses under the model presented in Section~\ref{sec:paqs} and use our algorithm (see Section~\ref{sec:estimator}) to generate a metric estimate. To simplify the example, all queries responses are generated in a \emph{noiseless} fashion---for example, the triplet comparison always returns the closer item to the reference.

We present our results in Figure~\ref{fig:noiseless_sims}, which illustrates a significant gap in information richness between \queryabbrs and a variety of ordinal queries. The number of \queryabbr responses needed to attain a reasonable normalized error levels is dramatically lower than those of typical ordinal queries. For example, to achieve a normalized error of 0.2, one needs at minimum 1,000 of any of the ordinal queries but only approximately $600$ PAQ responses. Overall, Figure~\ref{fig:noiseless_sims} quantitatively illustrates that \queryabbrs can greatly improve upon the performance of existing ordinal queries on metric learning. The rest of our paper explores this opportunity: It aims to make the deployment of \queryabbrs theoretically grounded by designing provable methodology for learning a low-rank metric from \queryabbr responses.

\subsection{Our contributions and organization}
\label{sec:intro:contributions}
In addition to introducing the \emph{\queryname} (\queryabbr), we demonstrate its applicability to metric learning under a Mahalanobis metric. We first present a mathematical formulation of this estimation problem in Section~\ref{sec:paqs}.
We then show that the sliding bar response can be viewed as an \emph{inverted measurement} of the metric matrix that we want to estimate,
which allows us to restate our problem as that of estimating a low-rank matrix from a specific type of trace measurement (Section~\ref{sec:estimator}).
However, our \queryabbr formulation differs from classical matrix estimation due to two technical challenges: (a) the sensing matrices and noise are correlated, and (b) the sensing matrices are heavy-tailed. As a result, standard matrix estimation algorithms give rise to \emph{biased estimators}. We propose a query procedure and an estimator that overcome these two challenges, and we prove statistical error bounds on the estimation error (Section~\ref{sec:main_results}). The unconventional nature of the sensing model and estimator causes unexpected behaviors in our error bounds; in Section~\ref{sec:experiments},
we present simulations verifying that these behaviors also appear in practice.

\subsection{Related work}
We now discuss related work in metric learning, along with the statistical techniques we use for our algorithm and analysis.
\paragraph{Metric learning.} In metric learning~\cite{bellet2022metric}, prior work considers using paired comparisons (of the form ``are these two items similar or dissimilar?'')~\cite{ying2009sparse, bian2012constrained, guo2014guaranteed, bellet2015robustness} and triplet comparisons (of the form ``which of the two items $\vx_1$ and $\vx_2$ is more similar to the reference item $\vx_0$?'')~\cite{mason2017learning}. 
The metric learning from triplets problem is generalized by~\cite{xu2020simultaneous} to consider an \emph{unknown} reference point (referred to as an ``ideal point'') that captures different individual preferences. Sample complexity guarantees for simultaneous estimation of a metric and individual ideal points are established in~\cite{canal2022one}. Tuple queries~\cite{canal2020tuplewise} extend triplets to ranking more than two items with respect to a reference item. The \queryabbr can be viewed as extending this set of items to a continuous spectrum, which is natural when one uses a generative model such as a GAN~\cite{goodfellow2014generative,karras2020analyzing}. However, the goal of tuple queries is to rank the items, whereas in \queryabbr the ranking is provided by the feature space and we ask people to identify a transition point (similar vs.\ dissimilar) in this ranking.

\paragraph{Statistical techniques.}
In our theoretical results, we apply techniques from the high-dimensional statistics literature. Our theoretical formulation (presented in Section~\ref{sec:estimator}) resembles the problem of low-rank matrix estimation from trace measurements (e.g.,~\cite{recht2010guaranteed,negahban2011estimation, tsybakov2011estimation, candes2011tight, negahban2012unified, cai2013sparse}; see \cite{davenport2016overview} for a more complete overview), and in particular, when the sensing matrix is of rank one and random~\cite{cai2015rop, chen2015exact, kueng2017low, mcrae2022optimal,chandrasekher2022alternating}. However, as discussed in Section~\ref{sec:estimator}, our model results in two important departures from prior literature. In our case, the sensing matrices are both heavy-tailed and correlated with the measurement noise, and the latter issue results in estimation bias for standard matrix estimation procedures. In addition, our heavy-tailed matrices violate the assumptions of much prior work that relies on sub-Gaussian or sub-exponential assumptions on the sensing matrices.
Prior work has attempted to address the challenge of heavy tails with methods such as robust loss functions \cite{loh2017statistical, fan2017estimation} or the ``median-of-means'' approach~\cite{nemirovskij1983problem, minsker2015geometric, hsu2016loss}, which partitions the data, constructs an estimator for each partition, and then forms one estimator based on some robustness criteria. We draw particular inspiration from~\cite{fan2021shrinkage}, which applies truncation to control heavy-tailed behavior in a number of problem settings. However, in the low-rank matrix estimation setting, the paper~\cite{fan2021shrinkage} only analyzes the case of heavy-tailed noise under a sub-Gaussian design, meaning that their methodology and results are not applicable to our problem setting.

\subsection{Notation}
For two real numbers $a$ and $b$, let $a \wedge b = \min\{a, b\}$ and $a \vee b = \max\{a, b\}$. Given a vector $\vx \in \bbR^d$, denote $\norm{\vx}_1$ and $\norm{\vx}_2$ as the $\ell_1$ and $\ell_2$ norm, respectively. Denote $\sphere^{d-1}\defn \{x\in \reals^d: \normtwo{x} = 1\}$ to be the set of $d$-dimensional vectors with unit $\ell_2$ norm. Given a matrix $\mA \in \bbR^{d_1 \times d_2}$, denote $\fronorm{\mA}$, $\nucnorm{\mA}$, and $\opnorm{\mA}$ as its Frobenius norm, nuclear norm, and operator norm, respectively. We denote $\bbS^{d \times d} = \{\mA \in \bbR^{d \times d} : \mA = \mA^\top\}$ to be the set of symmetric $d \times d$ matrices. Denote $\mA \succeq \vzero$ to mean that $\mA$ is symmetric positive semidefinite. For $\mA \succeq \vzero$, define the (pseudo-) norm $\norm{\vx}_{\mA} = \sqrt{\vx^\top\mA\vx}$. For matrices $\mA, \mB \in \bbR^{d_1 \times d_2}$, denote by $\ip*{\mA}{\mB} = \tr{\mA^\top\mB}$ the Frobenius inner product. For two sequences indexed by $x$, we use the notation $f(x) \lessorder g(x)$ to mean that there exists some absolute positive constant $\const > 0$, such that $f(x) \le \const\cdot g(x)$ for all $x$. We use the notation $f(x) \gtrorder g(x)$ when $g(x) \lesssim f(x)$.

\section{Formal model}\label{sec:paqs}
\begin{figure}
    \centering
    \includegraphics[width=0.7\textwidth]{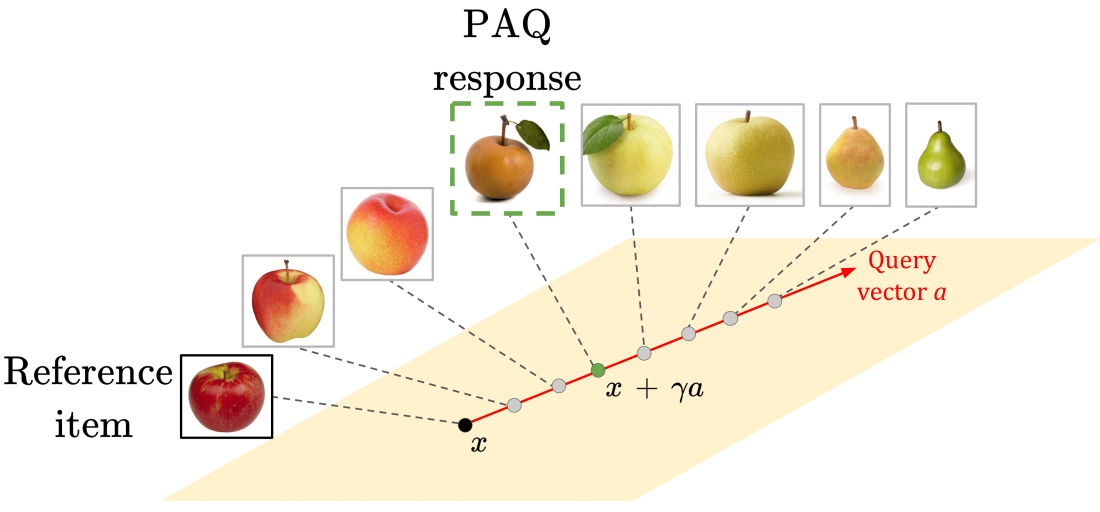}
    \caption{The \queryname. Given a reference item $\vx$ and a query vector $\va$, a continuous path of items is formed $\{\vx + \scale\va : \scale \in [0, \infty)\}$. Then, a user is asked to pick the first item along this path that is dissimilar to the reference item, denoted by $\vx + \scale\va$.}
    \label{fig:paq}
\end{figure}

In this section, we present our model for the \queryname (\queryabbr) in the context of its application to metric learning.

\subsection{Mahalanobis metric learning}\label{sec:paq:model}
We consider a $\dimension$-dimensional feature space where each item is represented by a point in $\reals^\dimension$. The distance metric model for human similarity perception posits that there is a metric on $\reals^\dimension$ that measures how dissimilar items are perceived to be.
A recent line of work \cite{xu2020simultaneous, canal2022one} has modeled the distance metric as a Mahalanobis metric.
If $\Sigstar \in \bbR^{d \times d}$ is a symmetric positive semidefinite (PSD) matrix,
the squared Mahalanobis distance with respect to $\Sigstar$ between items $\vx$ and $\vx' \in \reals^\dimension$ is $\normsigstar{\vx - \vx'}^2 \defn (\vx - \vx')^\top \Sigstar (\vx - \vx')$.
The distance represents the extent of dissimilarity between items $\vx$ and $\vx'$:
If we further have a perceptual boundary value $\boundary > 0$, this model posits that items $\vx, \vx'$ are perceived as similar if $\normsigstar{\vx - \vx'}^2 < \boundary$ and dissimilar if $\normsigstar{\vx -\vx'}^2 \geq \boundary$. We adopt a high-dimensional framework and, following existing work~\cite{mason2017learning, canal2022one}, assume that the matrix $\Sigstar$ is low-rank.

Note that if the goal is to predict whether two items are similar or dissimilar via computing the relation $\normsigstar{\vx - \vx'}^2 \gtrless \boundary$, then this problem is scale-invariant, in the sense that two items are predicted as similar (or dissimilar) according to $(\Sigstar, \boundary)$, if and only if they are predicted as similar (or dissimilar) according to $(\constscale\Sigstar, \constscale\boundary)$ for any scaling factor $\constscale> 0$. We are thus interested in finding the equivalence class of solutions $\{(\constscale \Sigstar, \constscale \boundary): \constscale > 0\}$.

\begin{remark}[Choice of $\boundary$]\label{rem:y-val} 
Since the goal is to learn (dis-)similarity between items,
one can set the boundary value to be any positive scalar $\boundary$, and estimate the matrix $\Sigstar$ corresponding to this value of $\boundary$. Indeed, our theoretical results proving error bounds on $\normfro{\Sighat - \Sigstar}$ exhibit a natural scale-equivariant property (see Section~\ref{sec:main_results}, Scale Equivariance).
\end{remark}

\subsection{The \queryname (\queryabbr)}\label{sec:paq:paq_def}
We assume that every point in our feature space $\reals^\dimension$ corresponds to some item. 
Recall from \Cref{fig:ui} that a
\queryabbr collects similarity data between a pair of items, where a reference item is fixed, and a spectrum of target items is generated from a one-dimensional path in the feature space.
Denote the reference item by $\vx\in \reals^\dimension$. The target items can be generated by any path in $\reals^\dimension$,
but for simplicity, we consider straight lines.
For any vector $\va\in \reals^\dimension$, we construct the line $\{\vx + \scale \va : \scale \in [0, \infty)\}$. We call this vector $\va$ the \emph{query vector}.
As shown in Figure~\ref{fig:paq}, the user moves the slider from left to right, and the value of $\scale$ increases proportionally to the distance traversed by the slider. Note that the value $\scale$ is \emph{dimensionless}.

The user is instructed to stop the slider at the transition point where the target item transitions from being similar to dissimilar with the reference item. According to our model, this transition point occurs when the $\Sigstar$-Mahalanobis distance between the target item and the reference item is $\boundary$. The (noiseless) transition point, denoted by $\scalestar$, satisfies the equation
\begin{align}
    \label{eq:scale_eq_ideal}
    \boundary = \normsigstar{\vx - (\vx + \scalestar\va)}^2 = \scalestar^2\quadSig{}.
\end{align}
Note that the ideal \queryabbr response $\scalestar$ does not depend on the specific reference item $\vx$ but rather only on the query direction $\va$ and the (unknown) metric matrix $\Sigstar$.
When querying users with \queryabbrs, the practitioner has control over how the query vectors $\va$ are selected. We discuss how to select $\va$ in Section~\ref{sec:estimator:algorithm}.

\subsection{Noise model}\label{sec:paq:noise_model}
We model the noise in human responses as follows: In the \queryabbr response~\eqref{eq:scale_eq_ideal}, we replace the boundary value $\boundary$ by $\boundary + \noise$, where $\noise \in \reals$ represents noise.
Thus the user provides a noisy response $\scale$ whose value satisfies $\scale^2\quadSig{} = \boundary + \noise$. Substituting in~\eqref{eq:scale_eq_ideal}, we have
\begin{align}
    \scale^2 = \scalestar^2 + \frac{\noise}{\quadSig{}}.\label{eq:noise_model}
\end{align}
If $\va^\top \Sigstar \va$ is large, then in the user interface Figure~\ref{fig:ui} (bottom), the semantic meaning of the item changes rapidly as the user moves the slider along the direction $\va$, and the slider stops at a position that is close to the true transition point. On the other hand, if $\va^\top \Sigstar \va$ is small, then the image changes slowly as the user moves the slider. It is hard to distinguish where exactly the transition occurs, so the slider ends up in a larger interval around the transition point. Recall that the scaling $\scale$ is proportional to the distance traversed by the slider.
This model~\eqref{eq:noise_model} thus captures such variation in the noise level, where the noise term $\frac{\noise}{\quadSig{}}$ is small when $\quadSig{}$ is large, and vice versa.

\section{Methodology}\label{sec:estimator} 
In this section, we formally present the statistical estimation problem for metric learning from noisy \queryabbr data, and we develop our algorithm for estimating the true metric matrix $\Sigstar$.

\subsection{Statistical estimation}
Assume we collect $N$ \queryabbr responses, using $N$ query vectors $\{\va_\idxitem\}_{\idxitem=1}^N$ that we select\footnote{
    In the sequel, we use the terms ``responses''/``measurements'' interchangeably for $\scale$, and the terms ``query vector''/``sensing vector'' interchangeably for $\va$.
}.
Denote the noise associated with these queries by random variables $\noise_1, \ldots, \noise_N\in \reals$. We obtain \queryabbr responses, denoted by $\scale_1, \ldots, \scale_N$, that satisfy  
\begin{align}
    \scale_\idxitem^2 \va_\idxitem^\top \Sigstar \va_\idxitem = \boundary + \noise_\idxitem, ~~ \idxitem = 1, \dots, N.\label{eq:paq_def}
\end{align}
We assume the noise variable $\noise$ is independent%
\footnote{This could be relaxed by placing conditions on the \emph{conditional} distributions of $\eta$ given $\va$ (and even the reference point $\vx$), but we omit this for simplicity.}%
of the query $\va$,
has zero mean and variance $\noisevar$, and is bounded, with $-y \leq \noise \leq \noiseup$ for some constant $\noiseup \geq 0$.
Note that we must have $\eta + y \geq 0$ since $\scale^2 \geq 0$; in addition, we place an upper bound $\noiseup$ on the noise.

Given the query directions $\{\va_\idxitem\}_{\idxitem=1}^N$ and the \queryabbr responses $\{\scale_\idxitem\}_{\idxitem=1}^N$, we want to estimate the matrix $\Sigstar$.
We first rewrite our measurement model as follows:
Recall that the matrix inner product is denoted by  $\ip{\mA}{\mB} \defn \tr{\mA^\top\mB}$ for any two matrices $\mA$ and $\mB$ of compatible dimension. Then from~\eqref{eq:paq_def}, we write
\begin{align}\label{eq:inverted_scaling}
    \scale^2 = \frac{\boundary + \noise}{\va^\top \Sigstar \va}.
\end{align}
Plugging~\eqref{eq:inverted_scaling} once more into \eqref{eq:paq_def}, we have
\begin{align*}
        \boundary + \noise = \ip{\mAinv}{\Sigstar},
\end{align*}
where
\begin{align}\label{eq:im_sensing_matrix}
    \mAinv \defn \scale^2\opa{} = \frac{\boundary + \noise}{\quadSig{}}\opa{}.
\end{align}
Hence, our problem resembles trace regression, and, in particular, low-rank matrix estimation from rank-one measurements (because the matrix $\mAinv$ has rank $1$)~\cite{cai2015rop,chen2015exact,kueng2017low,mcrae2022optimal}. We call $\mAinv$ the sensing matrix, and $\va$ the sensing vector.
Classical trace regression assumes that we make (noisy) observations of the form $y = \ip{\mA}{\Sigstar} + \epsilon$ where $\mA$ is fixed before we make the measurement; in our problem, the sensing matrix $\mAinv$ depends on our observed response $\scale$ and associated sensing vector $\va$.
Hence, the process of obtaining a \queryabbr response can be viewed as an \emph{inversion} of the standard trace measurement process. The inverse nature of our problem makes estimator design more challenging, as we discuss in the following section.

\subsection{Algorithm}\label{sec:estimator:algorithm}

As our first attempt at a procedure to estimate $\Sigstar$, we follow the literature~\cite{negahban2011estimation, mcrae2022optimal} and consider randomly sampling i.i.d.\ vectors $\va_\idxitem \sample \normal(\vzero, \mId_\dimension)$.
We then use standard least-squares estimation of $\Sigstar$.
Since we expect $\Sigstar$ to be low-rank, we add nuclear-norm regularization to promote low rank.
In particular, we solve the following program:
\begin{align}\label{eq:estimator_simplify}
    \min_{\mSigma \succeq \vzero}~\frac{1}{N} \sum\limits_{i=1}^{N} \left(\boundary - \ip{\mAinv_i}{\mSigma}\right)^2 + \lambda_N \nucnorm{\mSigma},
\end{align}
where $\lambda_N > 0$ is a regularization parameter.
This is a convex semidefinite program and can be solved with standard off-the-shelf solvers.

However, the inverted form of our measurement model creates two critical issues when naïvely using~\eqref{eq:estimator_simplify}:
\begin{itemize}[leftmargin=*]
    \item \textbf{Bias of standard matrix estimators due to dependence.}
    Note that the sensing matrix~\eqref{eq:im_sensing_matrix} depends on the noise $\noise$.
    Quantitatively, we have $\E{\noise \mAinv} \neq \vzero$ (see Appendix~\ref{app:prelim:not_cond_mean_zero}).
    Standard trace regression analyses require that this quantity be zero,
    typically assuming (at least) that $\noise$ is zero-mean conditioned on the sensing matrix $\mA$.
    The failure of this to hold in our case introduces a bias that does not decrease with the sample size $N$.

    \item \textbf{Heavy-tailed sensing matrix.}
    The factor $\frac{1}{\quadSig{}}$ in $\mAinv$ (see Equation~\eqref{eq:im_sensing_matrix}) makes $\mAinv$ heavy-tailed in general.
    When $\va$ is Gaussian, the term $\frac{1}{\quadSig{}}$ is an inverse weighted chi-square random variable, whose higher-order moments are infinite (and the number of finite moments depends on the rank of $\Sigstar$).
    This makes error analysis more difficult, as standard analyses require the sensing matrix $\mA$ to concentrate well (e.g., be sub-exponential).
\end{itemize}
To overcome these challenges, we make two key modifications to the procedure~\eqref{eq:estimator_simplify}.

\paragraph{Step 1: Bias reduction via averaging.}
First, we want to mitigate the bias due to the dependence between the sensing matrix $\mAinv$ and the noise $\noise$.
The bias term $\E{\noise \mAinv}$ scales proportionally to $\E{\noise(y + \noise) } = \E{\noise^2}$.
Therefore, to reduce this bias in the least-squares estimator \eqref{eq:estimator_simplify},
we need to reduce the noise variance.
We reduce the effective noise variance (and hence the bias) by \emph{averaging} i.i.d.\ samples.
Operationally, instead of obtaining $N$ measurements from $N$ distinct sensing vectors $\{\va_\idxitem\}_{\idxitem=1}^N$, we draw $n$ sensing vectors $\{\va_\idxitem\}_{\idxitem=1}^n$, and collect $m$ measurements, denoted by $\{\scale_\idxitem^{(j)}\}_{j=1}^m$ , corresponding to each sensing vector $\va_\idxitem$. We refer to $n$ as the number of (distinct) sensing vectors. To keep the total number of measurements constant, we set $n = \frac{N}{m}$, where the value of $m$ is specified later.
For each sensing vector $\va_\idxitem$, we compute the empirical mean of the $m$ measurements:
\begin{align}\label{eq:avg_measurement}
    \scalebar_i^2 \defn \frac1m \sum\limits_{j=1}^m (\scale_{i}^{(j)})^2
    = \frac{1}{m}\sum_{j=1}^m \frac{\boundary + \noise_\idxitem^{(j)}}{\quadSig{\idxitem}}
    = \frac{\boundary + \noisebar_i}{\quadSig{i}},
\end{align}
where we define the average noise by $\noisebar_i \defn  \frac1m\sum\limits_{j=1}^m \noise_{i}^{(j)}$. This averaging operation reduces the effective noise variance from $\var(\noise_\idxitem) = \varnoise$ to $\var(\noisebar_\idxitem) = \frac{\varnoise}{m}$.
If $n$ is small, we may have large error due to an insufficient number of query vectors $\va_\idxitem$.
On the other hand, a small $m$ leads to a large bias.
Therefore, we set the value of $m$ carefully to balance these two effects. This is studied theoretically in \Cref{sec:main_results} and demonstrated empirically in \Cref{sec:experiments}.

\paragraph{Step 2: Heavy tail mitigation via truncation.}
Next, we need to control the heavy-tailed behavior introduced by the $\frac{1}{\quadSig{}}$ term in the sensing matrix $\mAinv$.
Note that the sample averaging procedure \eqref{eq:avg_measurement} does not mitigate this problem.
We adopt the approach in \cite{fan2021shrinkage} and truncate the observations. Specifically, we truncate the averaged measurements $\scalebar_i^2$ by $\trunc$:
\begin{align}\label{eq:trunc_measurement}
    \scaletilde_i^2 \defn \scalebar_i^2 \wedge \trunc = \frac{\boundary + \noisebar_{\idxitem}}{\quadSig{\idxitem}} \wedge \tau,
\end{align}
where $\trunc > 0$ is a truncation threshold that we specify later. We then construct the truncated sensing matrices
\begin{align}\label{eq:truncated_sensing_matrix}
    \mAtilde_\idxitem = \scaletilde_{\idxitem}^2\opa{\idxitem} = \left(\frac{\boundary + \noisebar_{\idxitem}}{\quadSig{\idxitem}} \wedge \tau \right)\opa{\idxitem}, \quad i = 1, \dots, n.
\end{align}
While truncation mitigates heavy-tailed behavior, it also introduces additional bias in our estimate. The truncation threshold $\trunc$ therefore gives us another tradeoff, and in our analysis to follow, we carefully set the value of $\trunc$ to balance the effects of heavy-tailedness and bias.

\begin{algorithm}[t]
    \vspace{2mm}
	\caption{Inverted measurement sensing, averaging, and truncation.}
	\label{alg:estimator} 
 
    \textbf{Input:} number of total measurements $N$, averaging parameter $m$ (that divides $N$), truncation threshold $\trunc$, measurement value $\boundary$
    \vspace{1mm}
    \begin{algorithmic}[1]
		\STATE Compute the number of sensing vectors  $n = \frac{N}{m} $
		\FOR{$i = 1$ \textbf{to} $n$}
    		\STATE Draw sensing vector $\va_i$ from standard multivariate normal distribution
    		\STATE Obtain $m$ \queryabbr measurements $(\scale_{i}^{(1)})^2, \ldots, (\scale_{i}^{(m)})^2$ of the form
                \begin{align*}
                    (\scale_{i}^{(j)})^2 = \frac{\boundary + \noise_i^{(j)}}{\va_i^\top \Sigstar \va_i}, 
                \end{align*}
                where $\noise_i^{(j)}$ is an i.i.d. copy of the random noise $\noise$ for all $i$ and $j$
            \ENDFOR
            \FOR{$i = 1$ \textbf{to} $n$}
    		\STATE \label{line:averaging} Bias elimination via averaging: compute averaged response
            \begin{align*}
                \scalebar_i^2 = \frac1m\sum\limits_{j=1}^m (\scale_{i}^{(j)})^2.
            \end{align*}
    		\STATE \label{line:truncation} Heavy tail mitigation via truncation: compute truncated response 
            \begin{align*}
                \scaletilde_i^2 = \scalebar_i^2 \wedge \trunc.
            \end{align*}
		\ENDFOR
	\end{algorithmic}
    \vspace{1mm}
\textbf{Output:} truncated responses $\scaletilde_1^2, \ldots \scaletilde_n^2$
\vspace{2mm}
\end{algorithm}
\vspace{3mm}

\paragraph{Final algorithm.} Before presenting our final optimization program, we summarize our assumptions and sensing model below.

\begin{assumption}[Zero-mean, bounded noise]\label{assumption:noise}
    The observed noise values $\noise_i$ are i.i.d copies of the random variable $\noise$, which is independent of the random sensing vector $\va$. The random noise satisfies
    \begin{itemize}
        \item Zero-mean: $\E{\noise} = 0$
        \item Bounded: There exists a positive constant $\noiseup$ such that $-\boundary \leq \noise \leq \noiseup$ with probability $1$.
    \end{itemize}
\end{assumption}

We choose the sensing vector distribution to be the standard multivariate normal distribution and collect, average, and truncate $N$ PAQ responses following~\Cref{alg:estimator}. This process yields $n$ truncated responses $\scaletilde_1^2, \ldots \scaletilde_n^2$. We then use these truncated responses to form the averaged and truncated matrices $\{\mAtilde_\idxitem\}_{i=1}^n$, which we substitute into the original least-squares problem \eqref{eq:estimator_simplify}. To estimate $\Sigstar$, we solve
\begin{align}\label{eq:estimator}
    \Sighat \in \argmin_{\mSigma \succeq \vzero}~\frac{1}{n} \sum\limits_{i=1}^{n} \left(\boundary - \ip{\mAtilde_\idxitem}{\mSigma}\right)^2 + \lambda_n \nucnorm{\mSigma},
\end{align}
where, again, $\lambda_n$ is a regularization parameter that we specify later.

\paragraph{Practical considerations.}
In the averaging step, we collect $m$ measurements for each sensing vector $\va_\idxitem$. These measurements could be collected from $m$ different users. Furthermore, recall from Section~\ref{sec:paq:paq_def} that the measurements do not depend on the reference item $\vx$. As a result, one may also collect multiple responses from the same user by presenting them the same query vector $\va_i$ but with different reference items $\vx$. 
In addition, recall from Section~\ref{sec:paq:model} that user responses are scale-invariant. Practitioners are hence free to set the boundary $\boundary$ to be any positive value of their choice without loss of generality, and the noise variance $\varnoise$ scales accordingly with $\boundary$. The user interface does not depend on the value of $\boundary$.

\section{Theoretical results}\label{sec:main_results}
We now present our main theoretical result, which is a finite-sample error bound for estimating a low-rank metric from inverted measurements with the nuclear norm regularized estimator~\eqref{eq:estimator}.
Our error bound is generally stated, and depends on the averaging parameter $m$ and the truncation threshold $\trunc$.

Recall that $\noisevar$ denotes the variance of $\noise$. We define the quantities $\boundaryup \defn \boundary + \noiseup$ and $\noisemedian = \boundary + \texttt{median}(\noise)$. We further denote by $\sigma_1 \geq \cdots \geq \sigma_r > 0$ the non-zero singular values of $\Sigstar$.

\begin{theorem}\label{theorem:upper_bounds}
    Suppose $\Sigstar$ is rank $r$, with $r > 8$. Assume that we choose the sensing vector distribution the be the standard multivariate normal distribution, that~\Cref{assumption:noise} holds on the noise, and that we collect, average, and truncate measurements following~\Cref{alg:estimator}. Further, assume that the truncation threshold $\trunc$ satisfies $\trunc \geq \frac{\noisemedian}{\tr{\Sigstar}}$. Then there are positive constants $c, C, C_1$, and $C_2$, such that if the regularization parameter and the number of sensing vectors satisfy
    \begin{align}\label{eq:reg_n_conditions}
    \lambda_n \ge C_1\left[\boundaryup \left(\frac{\boundaryup}{\sigma_r r}\sqrt{\frac{d}{n}} + \frac{d}{n}\tau + \left(\frac{\boundaryup}{\sigma_r r}\right)^2\frac{1}{\tau}\right) + \frac{1}{\sigma_r r}\frac{\noisevar}{m}\right] 
    \quad \text{ and } \quad n \geq C_2 rd,
    \end{align}
    then any solution $\Sighat$ to the optimization program~\eqref{eq:estimator} satisfies
    \begin{align}\label{eq:error_bound}
    	\fronorm{\Sighat - \mSigma^\star} \leq C \left(\frac{\tr{\Sigstar}}{\noisemedian}\right)^2\sqrt{r} \lambda_n
    \end{align}
    with probability at least $1 - 4\exp\left(-d\right) - \exp\left(-cn\right)$.
\end{theorem}
The proof of Theorem~\ref{theorem:upper_bounds} is presented in Section~\ref{sec:thm1_proof}. The two sources of bias discussed in \Cref{sec:estimator:algorithm} appear in the expression \eqref{eq:reg_n_conditions} for the regularization parameter $\lambda_n$ (and consequently in the error bound~\eqref{eq:error_bound}). The term scaling as $1/\trunc$ corresponds to the bias induced by truncation, and decreases as the truncation gets milder (i.e., as the threshold $\trunc$ gets larger). 
The term scaling as $\noisevar/m$ corresponds to the bias arising from dependence between the noise and sensing matrix. As discussed in \Cref{sec:estimator:algorithm}, in this model, $m$-averaging results in a bias that scales like $1/m$. 

Given the dependence of the estimation error bound on the parameters $m$ and $\trunc$, we carefully set these parameters to obtain a tight bound as a function of the number of \emph{total measurements} $N =mn$.
These choices for $m$ and $\trunc$, along with the final estimation error, are presented below in Corollary~\ref{cor:upper_bounds_rate}. 

\begin{corollary}\label{cor:upper_bounds_rate}
    Recall that $N = mn$. Assume that the conditions of Theorem~\ref{theorem:upper_bounds} hold, and set the values of the constants
    $(c, C, C_1, C_2)$ according to Theorem~\ref{theorem:upper_bounds}. Suppose that the number of total measurements satisfies
    \begin{align}\label{eq:N_conditions}
         N \geq \left\{2C_2^{\nicefrac{3}{2}}\frac{\noisevar}{(\boundaryup)^2}r^{\nicefrac{3}{2}}d\right\} \vee \Bigg\{C_2rd \Bigg\}.
    \end{align}
    Set the averaging parameter $m$ and truncation threshold $\trunc$ to be
    \begin{align}\label{eq:trunc_averaging_choices}
        m = \Bigg\lceil \left(\frac{\noisevar}{(\boundaryup)^2}\right)^{\nicefrac{2}{3}}\left(\frac{N}{d}\right)^{\nicefrac13} \Bigg\rceil \quad &\text{ and } \quad \trunc = \frac{\boundaryup}{\sigma_r r}\sqrt{\frac{N}{md}},
    \end{align}
    and set $\lambda_n$ equal to its lower bound in~\eqref{eq:reg_n_conditions}.
    With probability at least $1 - 4\exp(-d) - \exp\left(-cN/m\right)$, we have:
    
    \begin{enumerate}[label=(\alph*)]
        \item If 
        $
            \frac{\noisevar}{(\boundaryup)^2} > \sqrt{\frac{d}{N}},
        $
        then any solution $\Sighat$ to the optimization program~\eqref{eq:estimator} satisfies
        \begin{align}\label{eq:rate}
            \fronorm{\Sighat - \Sigstar} \leq C^\prime\ \frac{\sigma_1^2}{\sigma_r} \frac{(\boundaryup)^{\nicefrac43} (\noisevar)^{\nicefrac13}}{\noisemedian^2}\ r^{\nicefrac{3}{2}}\left(\frac{d}{N} \right)^{\nicefrac{1}{3}}.
        \end{align}
        \item If
        $
            \frac{\noisevar}{(\boundaryup)^2} \leq \sqrt{\frac{d}{N}},
        $
        then any solution $\Sighat$ to the optimization program~\eqref{eq:estimator} satisfies
        \begin{align}\label{eq:rate-sqrt}
            \fronorm{\Sighat - \Sigstar} \leq C^\prime\ \frac{\sigma_1^2}{\sigma_r} \left(\frac{\boundaryup}{\noisemedian}\right)^2r^{\nicefrac{3}{2}}\left(\frac{d}{N}\right)^{\nicefrac{1}{2}}.
        \end{align}
    \end{enumerate}
In both cases, $C^\prime = 3C \cdot C_1$.
\end{corollary}

The proof of Corollary~\ref{cor:upper_bounds_rate} is provided in~\Cref{sec:cor1_proof}. A few remarks are warranted about our error bounds~\eqref{eq:rate} and~\eqref{eq:rate-sqrt}.

\paragraph{Error rates and noise regimes.} Under the standard trace measurement model, it is known that if the measurement matrices are i.i.d.\ according to some sub-Gaussian distribution
and the number of measurements satisfies $N \gtrsim rd$, then nuclear norm regularized estimators achieve an error that scales like $\sqrt{\frac{rd}{N}}$(e.g., \cite{negahban2011estimation, tsybakov2011estimation}). Such a result is also known to be minimax optimal \cite{tsybakov2011estimation}. Allowing heavier-tailed assumptions on the sensing matrices, such as sub-exponential \cite{kueng2017low, negahban2012restricted} or bounded fourth moment \cite{fan2021shrinkage}, typically results in additional $\log d$ factors but does not impact the exponent $1/2$ in the error rate.
However, a crucial assumption in these results is that $\E{\noise \mAinv} = \vzero$, and thus there is no bias due to measurement noise. Our inverted measurement sensing matrix 
is not only heavy-tailed but also leads to bias (see Lemma~\ref{lemma:not_conditional_mean_zero} in Appendix~\ref{app:prelim:not_cond_mean_zero}). Nevertheless, we are able to reduce the bias and trade it for variance, ensuring consistent estimation in all regimes.

In Corollary~\ref{cor:upper_bounds_rate}, there are two distinct cases for error rate which correspond to two different noise regimes induced by the quantity $ \frac{\noisevar}{(\boundaryup)^2}$, which captures the noise level in our measurements. In particular, the two cases in Corollary~\ref{cor:upper_bounds_rate} correspond to two regimes with distinct bias behavior:
\begin{enumerate}[label=(\alph*)]
    \item High-noise regime: In this setting, the bias due to measurement noise is non-negligible. As a result, we employ averaging with large $m$, which results in the rate scaling as $(d/N)^{\nicefrac{1}{3}}$.
    \item Low-noise regime: In this setting, the measurement noise bias is dominated by the variance, and thus has negligible impact on the estimation error. As a result, we are able to achieve a rate of order $(d/N)^{\nicefrac{1}{2}}$, which is consistent with established results for low-rank matrix estimation.
\end{enumerate}

\paragraph{Sample complexity.}
Since the degrees of freedom in a rank-$r$ matrix of size $d \times d$ is of order $rd$, one expects that the minimum number of measurements to identify a rank-$r$ matrix is of order $rd$. This is reflected in \Cref{theorem:upper_bounds},
which assumes that the number of \emph{distinct} sensing vectors $\{\va_i\}$ satisfies $n \gtrsim rd$. In the high-noise regime, from~\eqref{eq:trunc_averaging_choices} in Corollary~\ref{cor:upper_bounds_rate}, we have that $m$ scales like $(N/d)^{\nicefrac{1}{3}}$. Thus, the total number of measurements is $N = mn \gtrorder (N/d)^{\nicefrac{1}{3}} \cdot rd \gtrorder N^{\nicefrac{1}{3}} d^{\nicefrac{2}{3}}r$, and hence $N \gtrorder r^{\nicefrac{3}
{2}}d$.
Given that the rank is assumed to be relatively small compared to the dimension, the extra factor of $\sqrt{\rank}$ is a relatively small price to pay to obtain consistent estimation. In the low-noise regime, it can be verified that $m = 1$ in~\eqref{eq:trunc_averaging_choices} due to the low-noise condition $\frac{\noisevar}{(\boundaryup)^2} \leq \sqrt{\frac{d}{N}}$. No averaging is needed, and we only require  $N =n \gtrsim rd$.

\paragraph{Dependence on rank.}  When compared to standard results, Corollary~\ref{cor:upper_bounds_rate} differs in its dependence on rank. First, the matrix $\Sigstar$ is assumed to have rank $r > 8$. This prevents the term $\frac{1}{\quadSig{}}$ from making the sensing matrices so heavy-tailed that even truncation does not help. We empirically show that the assumption of $r > 8$ is necessary in \Cref{sec:experiments}. Second, there is an additional factor of $r$ in our rate for both noise regimes.
To interpret this, note that if $\Sigstar$ has non-zero singular values in a fixed range, then $\E{\quadSig{}} = \tr{\Sigstar} \approx r$.
Since the ``magnitude" of the sensing matrix $\mAinv$ is inversely proportional to $\quadSig{}$, increasing $r$ decreases the magnitude of $\mAinv$ and thus also (for a fixed noise level) the signal-to-noise ratio.

\paragraph{Scale equivariance}\label{rem:scale-inv}
As discussed in Section~\ref{sec:paq:model}, the metric learning from \queryabbrs problem aims to find an equivalence class $\{(\constscale \mSigma, \constscale \boundary): \constscale > 0\}$, and the ground-truth $\Sigstar$ is defined with respect to a particular choice of $\boundary$. Accordingly, our error bounds are scale-equivariant: If we instead replaced $\boundary$ with $\constscale\boundary$, the bounds~\eqref{eq:rate} and~\eqref{eq:rate-sqrt} would scale linearly in $\constscale$. This fact is precisely verified in Appendix~\ref{app:scale-eq} and relies on the fact that the noise also scales appropriately in $\constscale$. As alluded to in Remark~\ref{rem:y-val}, practitioners may simply set $\boundary$ to be \emph{any} positive number to estimate a metric that reflects item (dis-)similarity.

\section{Numerical simulations}\label{sec:experiments}
In this section, we provide numerical simulations investigating the effects of the various problem and estimation parameters.
For all results, we report the normalized estimation error $\fronorm{\Sighat - \Sigstar} / \fronorm{\Sigstar}$ averaged over 20 trials.
Shaded areas (sometimes not visible) represent standard error of the mean. 
For all experiments, we follow \cite{mason2017learning} and generate the ground-truth metric matrix as $\Sigstar = \frac{d}{\sqrt{r}}\mU\mU^\top$, where $\mU \in \bbR^{d \times r}$ is a randomly generated matrix with orthonormal columns. The noise $\noise$ is sampled from a uniform distribution on $[-\noiseup, \noiseup]$ (where $\noiseup \leq \boundary$). We set the regularization parameter, truncation threshold, and averaging parameter in a manner consistent with our theoretical results (see Eqs.~\eqref{eq:reg_n_conditions} and~\eqref{eq:trunc_averaging_choices}), cross-validating to choose the constant factors.
We solve the optimization problem using \texttt{cvxpy} \cite{diamond2016cvxpy,agrawal2018rewriting}. Code for all simulations is provided at \url{https://github.com/austinxu87/paq}.

\begin{figure}[t]
\centering
\hfill
\subcaptionbox{\label{fig:sim:vary_d}}[0.31\textwidth]{\includegraphics[width=\linewidth]{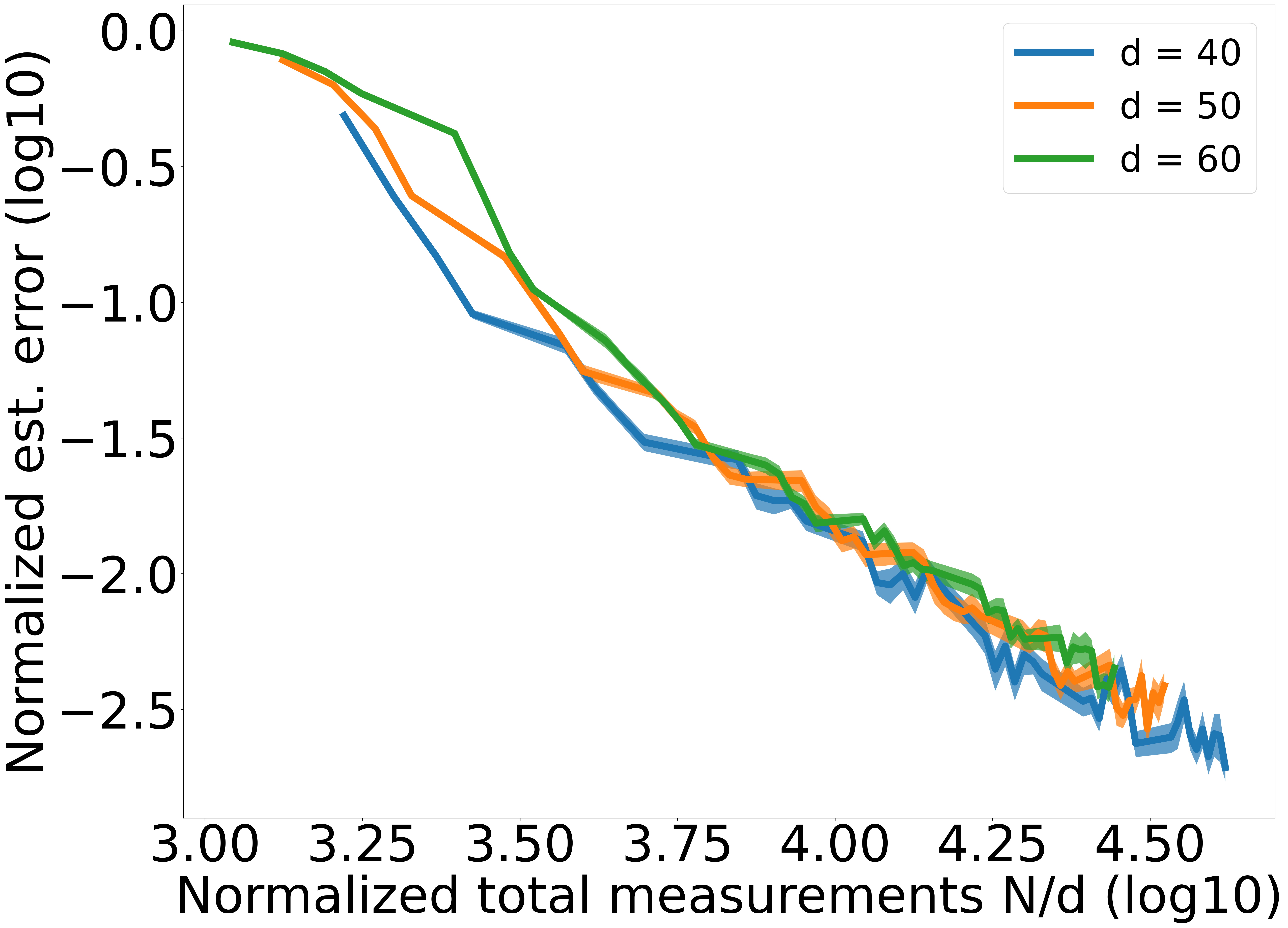}} 
\hfill
\subcaptionbox{\label{fig:sim:vary_r}}[0.31\textwidth]
{\includegraphics[width=\linewidth]{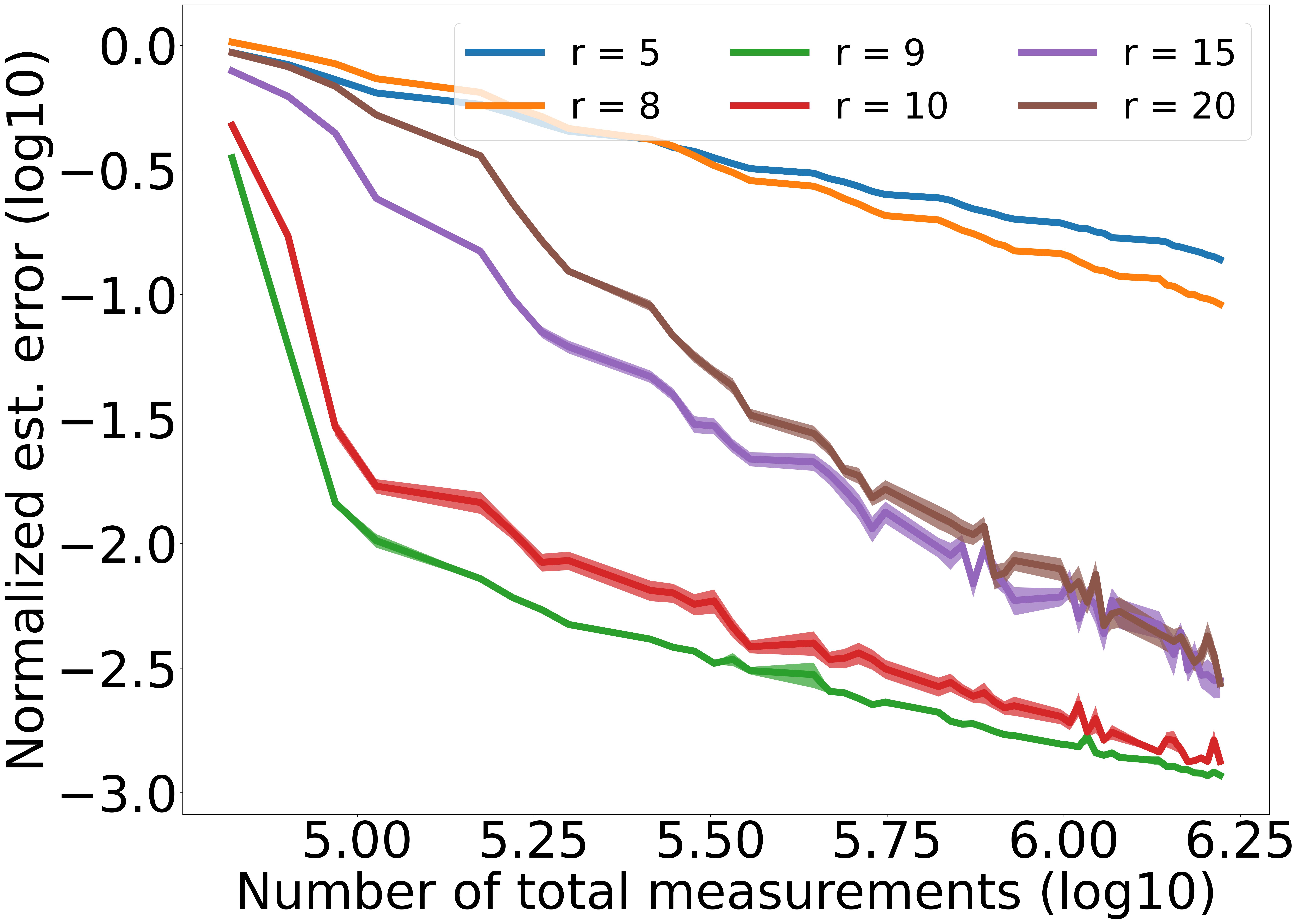}}
\hfill  
\subcaptionbox{\label{fig:sim:averaging}}[0.31\textwidth]{\includegraphics[width=\linewidth]{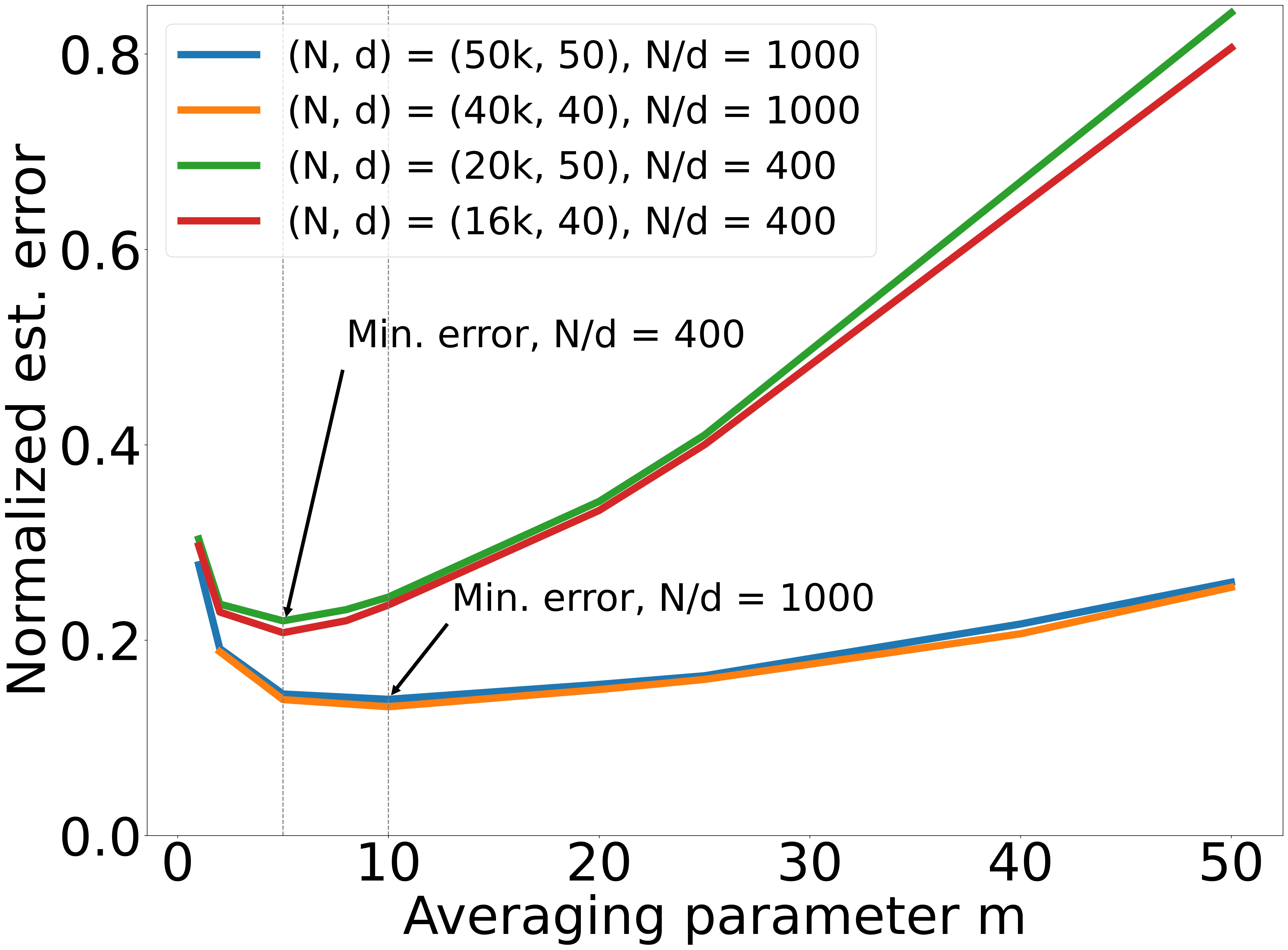}}
\hfill
\caption{Simulations quantifying the effect of dimension $d$, rank $r$, and averaging parameter $m$ on estimation error. Shaded areas correspond to standard error of the mean but sometimes not visible.}
\vspace{-2mm}
\end{figure}

\paragraph{Effects of dimension and rank.} Our first set of experiments characterizes the effects of dimension $d$ and matrix rank $r$. For all experiments, unless we are sweeping a specific parameter, we set $\boundary = 200$, $\dimension = 50$, $r = 15$, and $\noiseup = 10$. Fig.~\ref{fig:sim:vary_d} shows the performance for varying values of $\dimension$ plotted against the normalized sample size $N/d$. For all dimensions $\dimension$, the error decays to zero as the total number of measurements $N$ increases.
Furthermore, the error curves are well-aligned when the sample size is normalized by $\dimension$ with fixed $r$, empirically aligning with \Cref{cor:upper_bounds_rate}.
Fig.~\ref{fig:sim:vary_r} shows the performance for varying values of rank $\rank$. Recall that for our theoretical results we assume $\rank > 8$ to ensure that the quadratic term $\quadSig{}$ in the denominator of our sensing matrices does not lead to excessively heavy-tailed behavior.
When $r > 8$, the number of measurements required for the same estimation error increases as the rank increases. A clear phase transition occurs at $r = 8$. The error still decreases with $N$ for $r \leq 8$, but at a markedly slower rate than when $r > 8$.
This empirically demonstrates that when $r \leq 8$, the sensing matrix tails are too heavy to be mitigated by truncation. 

\paragraph{Effect of averaging parameter $m$.}  Equation~\eqref{eq:trunc_averaging_choices} suggests that the averaging parameter $m$ should scale proportionally to $(N/d)^{\nicefrac{1}{3}}$.
To test this, we set $\boundary = 200$, $d = 50$, $r = 9$, and $\noiseup = 200$.
We vary values of $m$ for different choices of the $(N, d)$ pair, as shown in Fig.~\ref{fig:sim:averaging}. The empirically optimal choice of $m$ is observed to be the same when $N/d$ is fixed, regardless of the particular choices of $N$ or $d$ (the green and red curves overlap, and the blue and orange curves overlap). Moreover, the optimal $m$ is smaller when $N/d = 400$  compared to when $N/d  = 1000$.

\section{Discussion}\label{sec:conclusion}
We introduce the \queryname, a cognitively lightweight way to obtain expressive human responses. We specifically investigate using \queryabbrs for human perceptual similarity learning. Learning models of human perception or preference has a range of applications, including recommendation systems, interrogating generative models, and quantifying perceptual conditions such as color blindness or hearing loss. We use a Mahalanobis distance-based model for human similarity perception and use \queryabbrs to estimate the unknown metric. This setup gives rise to a new inverted measurement scheme for high-dimensional low-rank matrix estimation which violates commonly held assumptions for existing estimators. We develop a two-stage estimator and provide corresponding sample complexity guarantees. 

This work lays the foundation for future work in two directions: (1) practical deployment of \queryabbrs and (2) theoretical characterization of learning from inverted measurements. One important aspect of deploying \queryabbrs in practice is how to select the most informative query directions. While this work considers a random query direction scheme that is amenable for theoretical analysis, targeted selection of query directions may reduce the number of responses needed in practice. Conducting user studies to collect data from human responses will also bring additional insights into how the theoretical guarantees translate into practice.

Along theoretical lines, one key direction is to characterize the optimal rate for this problem by deriving information-theoretic lower bounds. It is possible that there exists a fundamental trade-off between the variance and the bias that arises from the measurement scheme; it is also possible that more sophisticated techniques are capable of overcoming such bias.

\section{Proof of Theorem~\ref{theorem:upper_bounds}}\label{sec:thm1_proof}
Recall that we assume we collect $N$ measurements under the inverted measurements sensing model presented in~\Cref{alg:estimator} with standard Gaussian sensing vectors and bounded noise, mean-zero noise (\Cref{assumption:noise}). 

We first introduce a restricted strong convexity (RSC) condition that our proof relies on. Since the matrix $\Sigstar$ is assumed to be symmetric positive semidefinite matrices and of rank $\rank$, we follow~\cite{negahban2011estimation} and consider a restricted set on which we analyze the behavior of the sensing matrices $\mAtilde_\idxitem$. We call this set the ``error set'', defined by:
\begin{align}\label{eq:error_set}
\calE = \left\{ \mU \in \bbS^{d\times d} : \nucnorm{\mU} \leq 4\sqrt{2r}\fronorm{\mU} \right\},
\end{align}
where recall that $\bbS^{d\times d}$ denotes the set of symmetric $d \times d$ matrices.
We say that our shrunken sensing matrices $\{\mAtilde_\idxitem\}_{i=1}^n$ satisfy a restricted strong convexity (RSC) condition over the error set $\calE$, if there exists some positive constant $\kappa > 0$ such that
\begin{align}\label{eq:RSC_condition}
\frac{1}{n}\sum\limits_{i=1}^n\ip{\mAtilde_i}{\mU}^2 \geq \kappa\fronorm{\mU}^2 \qquad \text{ for all } \mU \in \calE.
\end{align}
The following proposition shows that the estimation error, when the sensing matrices satisfy the RSC condition and the regularization parameter is sufficiently large.
\begin{proposition}[{\cite[Theorem 1]{fan2021shrinkage}} with $q=0$]\label{prop:deterministic_error}
	Suppose that $\Sigstar$ has rank $r$ and the shrunken sensing matrices satisfy the restricted strong convexity condition~\eqref{eq:RSC_condition} with positive constant $\kappa >0$. Then if the regularization parameter satisfies
	\begin{align}\label{eq:op_norm}
	\lambda_n \geq 2\opnorm{\frac{1}{n}\sum\limits_{i=1}^n \boundary\mAtilde_i - \frac1{n}\sum\limits_{i=1}^n\ip{\mAtilde_i}{\Sigstar}\mAtilde_i  },
	\end{align} 
	any optimal solution $\Sighat$ of the optimization program~\eqref{eq:estimator} satisfies
	\begin{align}
	\fronorm{\Sighat - \Sigstar} \leq \frac{ 32\sqrt{r}\lambda_n }{ \kappa }. \nonumber
	\end{align}
\end{proposition}
This theorem is a special case of Theorem 1 in \cite{fan2021shrinkage}, which is in turn adapted from Theorem 1 in \cite{negahban2011estimation} (see~\cite{negahban2011estimation} or \cite{fan2021shrinkage} for the proof). Proposition~\ref{prop:deterministic_error} is a deterministic and nonasymptotic result and provides a roadmap for proving our desired upper bound. First, we show that the operator norm~\eqref{eq:op_norm} can be upper bounded with high probability, allowing us to set the regularization parameter $\lambda_n$ accordingly. Second, we show that the RSC condition~\eqref{eq:RSC_condition} is satisfied with high probability. We begin by bounding the operator norm~\eqref{eq:op_norm} in the following proposition.

\begin{proposition}\label{prop:operator_norm}
	Let $\boundaryup = \boundary + \noiseup$. Suppose that $\Sigstar$ has rank $r$, with $r > 8$.  Then there exists a positive absolute constant $C_1$ such that
	\begin{align}\label{eq:operator_norm}
		\opnorm{\frac{1}{n}\sum\limits_{i=1}^n \boundary\mAtilde_i - \frac1{n}\sum\limits_{i=1}^n\ip{\mAtilde_i}{\Sigstar}\mAtilde_i } \leq C_1\left[\boundaryup \left(\frac{\boundaryup}{\sigma_r r}\sqrt{\frac{d}{n}} + \frac{d}{n}\trunc + \left(\frac{\boundaryup}{\sigma_r r}\right)^2\frac1{\trunc}\right) + \frac{1}{\sigma_r r} \frac{\noisevar}{m}\right]
	\end{align}
	with probability at least $1 - 4\exp\left(-d\right)$. 
\end{proposition}
The proof of Proposition~\ref{prop:operator_norm} is provided in Section~\ref{sec:proof-op-norm}.
Next, we show that the RSC condition~\eqref{eq:RSC_condition} is satisfied with high probability, as is done in the following proposition.

\begin{proposition}\label{prop:rsc}
    Let $\noisemedian$ be the median of $\boundary + \noisebar$. Suppose that the truncation threshold $\trunc$ satisfies $\trunc \geq \frac{\noisemedian}{\tr{\Sigstar}}$. Then there exist positive absolute constants $\kappa_\calL$, $c$, and $C$ such that if the number of sensing vectors satisfy
    \begin{align}
        \quad n \geq C rd \nonumber
    \end{align}
    then we have
    \begin{align}\label{eq:prop_rsc_condition}
        \frac{1}{n} \sum\limits_{i=1}^n \ip{\mAtilde_i}{\mU}^2 \geq \kappa_\calL \left(\frac{ \noisemedian}{\tr{\Sigstar} }\right)^2 \fronorm{\mU}^2
    \end{align}
    simultaneously for all matrices $\mU \in \calE$ with probability at least $1 - \exp(-cn)$, where $\calE$ is the error set defined in Equation~\eqref{eq:error_set}.
\end{proposition}

The proof of Proposition~\ref{prop:rsc} is provided in Section~\ref{sec:proof-rsc}. We now utilize the results of Propositions~\ref{prop:deterministic_error},~\ref{prop:operator_norm} and~\ref{prop:rsc} to derive our final error bound. By Proposition~\ref{prop:operator_norm}, the operator norm~\eqref{eq:op_norm} can be upper bounded with high probability. We set the regularization parameter $\lambda_n$ to satisfy
\begin{equation}
\lambda_n \geq C_1\left[\boundaryup \left(\frac{\boundaryup}{\sigma_r r}\sqrt{\frac{d}{n}} + \frac{d}{n}\trunc + \left(\frac{\boundaryup}{\sigma_r r}\right)^2\frac1{\trunc}\right) + \frac{1}{\sigma_r r}\frac{\noisevar}{m}\right],\nonumber
\end{equation}
where $C_1$ is the constant in Proposition~\ref{prop:operator_norm}. Furthermore, by Proposition~\ref{prop:rsc}, we have that there exist universal constant $C_2 > 0$ such that if the number of sensing vectors satisfies $n \geq C_2 rd$, the RSC condition~\eqref{eq:RSC_condition} holds for constant $\kappa = \kappa_\calL \left(\frac{\noisemedian}{\tr{\Sigstar}}\right)^2$ with high probability. Taking a union bound, we have that \Cref{prop:operator_norm} and \Cref{prop:rsc} hold simultaneously with probability at least $1-4\exp(-\dimension) - \exp(-\const n)$. Invoking Proposition~\ref{prop:deterministic_error}, we have 
\begin{align}
\fronorm{\Sighat - \Sigstar} &\leq 32\sqrt{r}\cdot \frac{\lambda_n}{ \kappa_\calL \left(\frac{\noisemedian}{\tr{\Sigstar}}\right)^2}\nonumber \\
&\lessorder \left(\frac{\tr{\Sigstar}}{\noisemedian}\right)^2 \sqrt{r} \lambda_n\nonumber
\end{align}
with probability at least $1 - 4\exp(-d) - \exp(-cn)$, as desired. 

\subsection{Proof of Proposition~\ref{prop:operator_norm}}\label{sec:proof-op-norm}
In the proof, we decompose the operator norm
$\opnorm{\frac1{n}\sum\limits_{i=1}^n \boundary\mAtilde_i - \frac1{n}\sum\limits_{i=1}^n\ip{\mAtilde_i}{\Sigstar}\mAtilde_i }$
from~\eqref{eq:operator_norm} into individual terms and bound them separately. We define random matrices
\begin{align}\label{eq:avg_sensing_random_matrix}
    \mAbar = \scalebar^2\opa{} = \frac{\boundary + \noisebar}{\quadSig{}}\opa{}
\end{align}
and
\begin{align}\label{eq:trunc_sensing_random_matrix}
    \mAtilde = \scaletilde^2\opa{} = \left(\frac{\boundary + \noisebar}{\quadSig{}} \wedge \trunc \right)\opa{}
\end{align}
as the sensing matrix formed with the $m$-averaged responses $\scalebar$ and truncated responses $\scaletilde$, respectively.

\paragraph{Step 1: decompose the error into five terms.}
We begin by adding and subtracting multiple quantities as follows:
\begin{align}
\frac1{n}\sum\limits_{i=1}^n \boundary\mAtilde_i - \frac1{n}\sum\limits_{i=1}^n\ip{\mAtilde_i}{\Sigstar}\mAtilde_i &= \frac1{n}\sum\limits_{i=1}^n \boundary\mAtilde_i - \E{\boundary\mAtilde} + \E{\boundary\mAtilde} - \E{\boundary\mAbar}	\nonumber \\ 
&\quad + \E{\boundary\mAbar} - \E{\ip{\mAtilde}{\Sigstar}\mAtilde} + \E{\ip{\mAtilde}{\Sigstar}\mAtilde} - \frac1{n}\sum\limits_{i=1}^n\ip{\mAtilde_i}{\Sigstar}\mAtilde_i \nonumber\\
&\overset{(i)}{=} \frac1{n}\sum\limits_{i=1}^n \boundary\mAtilde_i - \E{\boundary\mAtilde} + \E{\boundary\mAtilde} - \E{\boundary\mAbar} \nonumber \\ 
&\quad + \E{\ip{\mAbar}{\Sigstar}\mAbar} - \E{\ip{\mAtilde}{\Sigstar}\mAtilde} - \E{\noisebar\mAbar} \nonumber \\ 
&\quad + \E{\ip{\mAtilde}{\Sigstar}\mAtilde} - \frac1{n}\sum\limits_{i=1}^n\ip{\mAtilde_i}{\Sigstar}\mAtilde_i,\label{eq:decompose_terms}
\end{align}
where step~\stepone is true by substituting $\boundary = \ip{\mAbar}{\Sigstar} - \noisebar$ to the term of $\E{\boundary\mAbar}$, and the fact that the noise term $\noisebar$ is zero-mean. By triangle inequality, we group the terms in~\eqref{eq:decompose_terms} and bound the operator norm by
\begin{align}
\opnorm{\frac1{n}\sum\limits_{i=1}^n \boundary\mAtilde_i - \frac1{n}\sum\limits_{i=1}^n\ip{\mAtilde_i}{\Sigstar}\mAtilde_i } &\leq \underbrace{\boundary\opnorm{\frac1{n}\sum\limits_{i=1}^n \mAtilde_i - \E{\mAtilde}}}_{\text{Term 1}} \nonumber \\ 
&+ \underbrace{\boundary\opnorm{\E{\mAtilde} - \E{\mAbar}}}_{\text{Term 2}} \nonumber \\
&+ \underbrace{\opnorm{\E{\ip{\mAbar}{\Sigstar}\mAbar} - \E{\ip{\mAtilde}{\Sigstar}\mAtilde}}}_{\text{Term 3}} \nonumber \\
&+ \underbrace{\opnorm{\E{\ip{\mAtilde}{\Sigstar}\mAtilde} - \frac1{n}\sum\limits_{i=1}^n\ip{\mAtilde_i}{\Sigstar}\mAtilde_i}}_{\text{Term 4}} \nonumber \\
&+ \underbrace{\opnorm{\E{\noisebar\mAbar}}}_{\text{Term 5}}.\label{eq:five_terms}
\end{align}
In the remaining proof, we bound the five terms in~\eqref{eq:five_terms} individually. We first discuss the nature of these five terms. 
\begin{itemize}
    \item \textbf{Terms 1 and 4:} These two terms characterize the difference between the empirical mean of quantities involving $\mAtilde$ and their true expectation (see \Cref{lemma:op_bound_1} and \Cref{lemma:op_bound_4}). In the proof, we show that the empirical mean concentrates around the expectation with high probability, as a function of the number of sensing vectors $n$.

    \item \textbf{Terms 2 and 3:} These two terms characterize the difference in expectation introduced by truncating $\mAbar$ to $\mAtilde$ (see \Cref{lemma:op_bound_2} and \Cref{lemma:op_bound_3}). Hence, these two terms characterize biases that arise from truncation. They diminish as $\thresh\rightarrow \infty$, because setting $\thresh$ to $\infty$ is equivalent to no thresholding, and hence $\mAtilde$ becomes identical to $\mAbar$. Since expectations are considered, these two terms depend on the threshold $\thresh$, but not the number of sensing vectors $n$ or the averaging parameter $m$. 

    \item \textbf{Term 5:} Term 5 is a bias term that arises from the fact that the mean of the noise $\eta$ conditioned on sensing matrix $\mAbar$ is non-zero. We show that this bias scales like $\frac1m$ (see \Cref{lemma:op_bound_5}) in terms of the averaging parameter $m$.
\end{itemize}
Putting these terms together, Terms 1 and 4 depend on $n$, Terms 2 and 3 depend on $\tau$, and Term 5 depends on $m$. In \Cref{cor:upper_bounds_rate}, we set the values of $\tau$, $n$ and $m$ to balance these terms.

\paragraph{Step 2: bound the five terms individually.} In what follows, we provide five lemmas to bound each of the five terms individually. In the proofs of the five lemmas, we rely on an upper bound on the fourth moment of the $m$-sample averaged measurements $\scalebar^2$. As shown in  Lemma~\ref{lemma:gamma_bar_fourth_moment} in Appendix~\ref{app:prelim:fourth_moment}, for some absolute constant $c$, this fourth moment can be upper bounded by a quantity that we denote $M$:
\begin{align}
    \Expect[(\scalebar^2)^4] \le \fourthmoment = c \left(\frac{\boundaryup}{\sigma_r r}\right)^4.\label{eq:fourth_moment_recall_prop2}
\end{align}
We also rely heavily on the following truncation properties relating the $m$-sample averaged measurements $\scalebar^2$ and truncated measurements $\tilde{\gamma}^2$:
\begin{align}
    & \tilde{\gamma}_i^2 \leq \trunc \label{tp1}\tag{TP1}\\
    & \tilde{\gamma}_i^2 \leq \scalebar_i^2 \label{tp2}\tag{TP2}\\
    & \scalebar_i^2 - \tilde{\gamma}_i^2 = (\scalebar_i^2 - \tilde{\gamma}_i^2)\cdot \vone\{\scalebar_i^2 \geq \trunc\}. \label{tp3}\tag{TP3}
\end{align}	
The following lemma provides a bound for Term 1.
\begin{lemma}\label{lemma:op_bound_1}
    Let $\mAtilde_1, \ldots, \mAtilde_n$ be i.i.d copies of a random matrix $\mAtilde$ as defined in Equation~\eqref{eq:trunc_sensing_random_matrix}. There exists an absolute constant $\const > 0$ such that for any $t > 0$, we have        
    \begin{align}
	\opnorm{\frac1{n}\sum\limits_{i=1}^n \mAtilde_i - \E{\mAtilde_i}} \leq  \const\left( \sqrt{\frac{\fourthmoment^{\nicefrac{1}{2}}t}{n}} + \frac{\trunc t}{n}\right)\nonumber 
    \end{align}
    with probability at least $1 - 2\cdot 9^d \cdot \exp\left(-t\right)$. 
\end{lemma}
The proof of Lemma~\ref{lemma:op_bound_1} is provided in Section~\ref{sec:op_bound_1}.
The next lemma provides an upper bound for Term 2.
\begin{lemma}\label{lemma:op_bound_2}
	Let $\mAbar$ and $\mAtilde$ be the random matrices defined in Equation~\eqref{eq:avg_sensing_random_matrix} and Equation~\eqref{eq:trunc_sensing_random_matrix}, respectively. Then there exists an absolute constant $\const > 0$ such that
	\begin{align}
	\opnorm{\E{\mAtilde} - \E{\mAbar}} \leq \frac{ \const \fourthmoment^{\nicefrac{1}{2}}}{\trunc}.\nonumber
	\end{align} 
\end{lemma}

The proof of Lemma~\ref{lemma:op_bound_2} is provided in Section~\ref{sec:op_bound_2}. The following lemma provides an upper bound for Term 3. Recall that the quantity $\boundaryup$ denotes $\boundary + \noiseup$.
\begin{lemma}\label{lemma:op_bound_3}
	Let $\mAbar$ and $\mAtilde$ be the random matrices defined in Equation~\eqref{eq:avg_sensing_random_matrix} and Equation~\eqref{eq:trunc_sensing_random_matrix}, respectively. Then there exists an absolute constant $\const > 0$ such that
	\begin{align}
	\opnorm{\E{\ip{\mAbar}{\Sigstar}\mAbar} - \E{\ip{\mAtilde}{\Sigstar}\mAtilde}} \leq \frac{ \const\ \boundaryup\fourthmoment^{\nicefrac{1}{2}}}{\trunc}.\nonumber
	\end{align}
\end{lemma}

The proof of Lemma~\ref{lemma:op_bound_3} is provided in Section~\ref{sec:op_bound_3}. The following lemma provides an upper bound for Term 4. 
\begin{lemma}\label{lemma:op_bound_4}
    Let $\mAtilde_1, \ldots, \mAtilde_n$ be i.i.d copies of a random matrix $\mAtilde$ defined in Equation~\eqref{eq:trunc_sensing_random_matrix}. There exists an absolute constant $\const > 0$ such that for any $t > 0$, we have        
    \begin{align}
            \opnorm{\E{\ip{\mAtilde}{\Sigstar}\mAtilde} - \frac1{n}\sum\limits_{i=1}^n\ip{\mAtilde_i}{\Sigstar}\mAtilde_i} \leq \const\ \boundaryup\left(\sqrt{\frac{\fourthmoment^{\nicefrac{1}{2}}t}{n}} + \frac{\trunc t}{n}\right)\nonumber
    \end{align}
    with probability at least $1 - 2\cdot 9^d \cdot \exp\left(-t\right)$. 
\end{lemma}
The proof of Lemma~\ref{lemma:op_bound_4} is provided in Section~\ref{sec:op_bound_4}.
We note that Terms 2 and 3 are bias that result from shrinkage, but crucially are inversely dependent on the shrinkage threshold $\trunc$. This fact allows us to set $\trunc$ so that the order of Terms 2 and 3 match the order of Terms 1 and 4.

The final lemma bounds Term 5, which is a bias that arises from the dependence of the sensing matrix $\mAbar$ on the noise $\noise$.
\begin{lemma}\label{lemma:op_bound_5}
    Let $\mAbar$ be the random matrix defined in Equation~\eqref{eq:avg_sensing_random_matrix}. Suppose that $\Sigstar$ has rank $r$ with $r > 2$. Then there exists an absolute constant $\const > 0$ such that
    \begin{align}
        \E{\opnorm{\noisebar\mAbar}} \leq \frac{\const}{\sigma_r r}\frac{\noisevar}{m}.\nonumber
    \end{align}
\end{lemma}
The proof of Lemma~\ref{lemma:op_bound_5} is provided in Section~\ref{sec:op_bound_5}. We note that the bias scales with the variance of the $m$-sample averaged noise $\noisebar$, which scales inversely with $m$. 

\paragraph{Step 3: combine the five terms.} 
We set $t = (\log9 + 1)d$. Substituting the bounds from Lemmas~\ref{lemma:op_bound_1}--~\ref{lemma:op_bound_5} back to~\eqref{eq:five_terms} and taking a union bound, we have that with probability at least $1 - 4\exp(-d)$, 
\begin{align}
    \opnorm{\frac1{n}\sum\limits_{i=1}^n \boundary\mAtilde_i - \frac1{n}\sum\limits_{i=1}^n\ip{\mAtilde_i}{\Sigstar}\mAtilde_i } &\lesssim \left(\boundaryup + 1\right)\left( \sqrt{\frac{\fourthmoment^{\nicefrac{1}{2}}d}{n} } + \frac{d}{n}\trunc + \frac{\fourthmoment^{\nicefrac{1}{2}}}{\trunc} \right) + \frac{1}{\sigma_r r}\frac{\noisevar}{m}\nonumber\\
    &\overset{(i)}{\lesssim} \boundaryup\left( \frac{\boundaryup}{\sigma_r r}\sqrt{\frac{d}{n}} + \frac{d}{n}\trunc + \left(\frac{\boundaryup}{\sigma_r r}\right)^2\frac{1}{\trunc}\right) + \frac{1}{\sigma_r r}\frac{\noisevar}{m},\nonumber
\end{align}

where step~\stepone is true by substituting in the expression~\eqref{eq:fourth_moment_recall_prop2} for $\fourthmoment$. 


\subsubsection{Proof of Lemma~\ref{lemma:op_bound_1}.}\label{sec:op_bound_1}
Let $\setcover_{\frac14}\subseteq \sphere^{d-1}$ be a $\frac14$-covering of the $d$-dimensional unit sphere $\sphere^{d-1}\defn \{x\in \reals^d: \normtwo{x} = 1\}$. By a covering argument~\cite[Exercise 4.4.3]{vershynin2018high}, for any symmetric matrix $\varmtx\in \bbS^{\dimension\times \dimension}$, its operator norm is bounded by $\normop{\varmtx} \le 2\sup_{\vv\in \setcover_\frac{1}{4}}\abs*{\vv^\top \varmtx \vv}$. Hence, we have
\begin{align}
\opnorm{\frac1{n}\sum\limits_{i=1}^n \mAtilde_i - \E{\mAtilde}} & \le 2 \sup_{\vv \in \setcover_{\frac14}} \left|\vv^\top\left(\frac1{n}\sum\limits_{i=1}^n \mAtilde_i - \E{\mAtilde}\right)\vv\right| \nonumber\\
& = 2\sup_{\vv \in \setcover_{\frac14}} \left|\frac1{n}\sum\limits_{i=1}^n \vv^\top\mAtilde_i\vv - \E{\vv^\top\mAtilde\vv} \right|.\label{eq:term_1_bound_cover}
\end{align}
We invoke Bernstein's inequality. We first show that
the Bernstein condition holds. Namely, we show that for each integer $\power \ge 2$, we have that for any unit vector $\vv \in \reals^\dimension$,
\begin{align}
    \Expect \;\abs*{\vv^\top \mAtilde \vv}^\power\le \frac{\power!}{2} \constbernone \constberntwo^{\power - 2},\label{eq:term_1_bern_condition}
\end{align}
where $\constbernone = \const_1\fourthmoment^\frac{1}{2} $ and $\constberntwo = \const_2 \tau$ for some universal positive constants $\const_1$ and $\const_2$. 
Given the Bernstein condition~\eqref{eq:term_1_bern_condition}, we then apply Bernstein's inequality to bound~\eqref{eq:term_1_bound_cover}.

\paragraph{Proving the Bernstein condition~\eqref{eq:term_1_bern_condition}.}
We fix any unit vector $\vv\in \reals^\dimension$.
Since $\mAtilde = \scaletilde^2 \opa{}$, we have
$\vv^\top\mAtilde\vv=\scaletilde^2(\vv^\top\va)^2$.
Recall that the random vector $\va$ is distributed as $\va \sim \calN(\vzero, \mId_d)$. Since $\vv$ is a unit vector, it follows that $\vv^\top\va \sim \calN(0, 1)$. Denote by $\vargauss\sample \normal(0, 1)$ a standard normal random variable. For any integer $\power \ge 2$, we have
\begin{align}
\Expect \; \abs*{\vv^\top \mAtilde \vv}^\power = \Expect\; \left(\scaletilde^2 \vargauss^2\right)^\power &\overset{\stepone}{\leq} \tau^{p-2} \E{\left(\scaletilde^2\right)^2\vargauss^{2\power}} \nonumber\\
&\overset{\steptwo}{\leq} \tau^{p-2} \cdot \Expect\left[\left(\scalebar^2\right)^2\vargauss^{2\power}\right] \nonumber\\
&\overset{\stepthree}{\leq} \tau^{p-2} \left(\E{\left(\scalebar^2\right)^4} \cdot \Expect\left[\vargauss^{4\power}\right]\right)^{\nicefrac{1}{2}}\nonumber\\
&\overset{\stepfour}{\leq} \tau^{p-2} \left(\fourthmoment\cdot \E{\vargauss^{4\power}}\right)^{\nicefrac{1}{2}}\label{eq:term1_bern_condition_expreession},
\end{align}
where steps~\stepone and~\steptwo follow from~\eqref{tp1} and~\eqref{tp2}, respectively; step~\stepthree follows from Cauchy--Schwarz inequality; and step~\stepfour follows upper bounding the fourth moment of $\scalebar^2$ with the quantity $\fourthmoment$ from Equation~\eqref{eq:fourth_moment_recall_prop2}.

Note that since $\vargauss$ is standard normal, by definition $\vargauss^2$ follows a Chi-Square distribution with $1$ degree of freedom, and hence sub-exponential. By Lemma~\ref{lemma:subexp_moments} in Appendix~\ref{app:prelim:sub_exp}, there exists some constant $\const > 0$ such that we have $\left(\E{(\vargauss^2)^\power}\right)^{\nicefrac{1}{\power}} \leq cp$ for all $\power \geq 1$. Hence, we have
$\left(\E{\vargauss^{4\power}}\right)^{\nicefrac{1}{2\power}} \leq 2cp$ and
\begin{align}
\left(\E{\vargauss^{4\power}} \right)^{\nicefrac{1}{2}} & \overset{}{\leq} \left(2cp\right)^p
= \left(\frac{p}{e}\right)^\power \cdot (2ec)^p\nonumber\\
& \stackrel{\stepone}{<} \power!\cdot (2ec)^\power\label{eq:term1_bern_condition_factorial}
\end{align}
where step~\stepone is true by Stirling's inequality that for all $p \ge 1$, 
\begin{align*}
    p! > \sqrt{2\pi p}\left(\frac{p}{e}\right)^p e^{\frac{1}{12p+1}} > \left(\frac{p}{e}\right)^\power.
\end{align*}
Substituting~\eqref{eq:term1_bern_condition_factorial} back to~\eqref{eq:term1_bern_condition_expreession} and rearranging terms completes the proof of the Bernstein condition~\eqref{eq:term_1_bern_condition}.

\paragraph{Applying Bernstein's inequality to bound~\eqref{eq:term_1_bound_cover}.}
By Bernstein's inequality (see Lemma~\ref{lemma:bernsteins_inequality}), given condition~\eqref{eq:term_1_bern_condition}, we have that for any unit vector $\vv\in \reals^\dimension$ and any $t > 0$,
\begin{align}
\P{\left|\frac1{n}\sum\limits_{i=1}^n \vv^\top\mAtilde_i\vv - \E{\vv^\top\mAtilde\vv}\right| \geq 2\left(\sqrt{\frac{\const_1\fourthmoment^{\nicefrac{1}{2}}t}{n}} + \frac{\const_2\tau t}{n}\right)} \leq 2\exp\left(-t\right).\label{eq:prop2_bernstein_inequality_any_v}
\end{align}
By~\cite[Corollary 4.2.13]{vershynin2018high}, the cardinality of the covering set $\setcover_{\frac14}$ is bounded above by $9^{\dimension}$. Therefore, taking a union bound on~\eqref{eq:prop2_bernstein_inequality_any_v}, we have
\begin{align}
\P{\sup_{\vv\in\setcover_{\frac14}}\left|\frac1{n}\sum\limits_{i=1}^n \vv^\top\mAtilde_i\vv - \E{\vv^\top\mAtilde\vv}\right| \geq 2\left(\sqrt{\frac{\const_1\fourthmoment^{\nicefrac{1}{2}}t}{n}} + \frac{\const_2\tau t}{n}\right) } \leq 2 \cdot 9^d \cdot \exp\left(-t\right).\label{eq:term1_union_bound}
\end{align}
Substituting in~\eqref{eq:term_1_bound_cover} to~\eqref{eq:term1_union_bound}, for any $t > 0$, we have
\begin{align}
\Prob\left(\opnorm{\frac1{n}\sum\limits_{i=1}^n \mAtilde_i - \E{\mAtilde}} \lesssim \sqrt{\frac{\fourthmoment^{\nicefrac{1}{2}}t}{n}} + \frac{\tau t}{n}\right)
\ge 1 -  2 \cdot 9^d \cdot \exp(-t), \nonumber
\end{align}
as desired.


\subsubsection{Proof of Lemma~\ref{lemma:op_bound_2}}\label{sec:op_bound_2}
By definition of the operator norm, we have
\begin{align}
\opnorm{\E{\mAtilde} - \E{\mAbar}} = \sup_{\vv \in \sphere^{d-1}} \abs*{\vv^\top \left( \E{\mAbar} - \E{\mAtilde} \right)\vv}\nonumber.
\end{align}
We fix any $\vv\in \sphere^{\dimension-1}$, and bound $\vv^\top \left( \E{\mAbar} - \E{\mAtilde} \right)\vv $. Similar to the proof of Lemma~\ref{lemma:op_bound_1}, we note that $\vv^\top\va \sim \normal(0, 1)$ and denote the random variable $\vargauss\sample \normal(0, 1)$. Substituting in the expression for sensing matrices $\mAbar$ and $\mAtilde$, we have
\begin{align}
    \abs*{\vv^\top \left( \E{\mAbar} - \E{\mAtilde} \right)\vv}
    &= \abs*{\vv^\top \E{\scalebar^2\opa{} - \scaletilde^2\opa{}}\vv}\nonumber\\
    & \stackrel{\stepone}{=} \E{\left(\scalebar^2 - \scaletilde^2\right)\vargauss^2}\nonumber\\
    &\overset{\steptwo}{=}  \E{\left(\scalebar^2 - \scaletilde^2\right)\vargauss^2\cdot \indicator\{\scalebar^2 \geq \tau\}}\nonumber\\
    &\overset{}{\leq} \E{\scalebar^2\vargauss^2\cdot \indicator\{\scalebar^2 \geq \tau\}}\nonumber\\
    &\overset{\stepthree}{\leq} \Big(\E{(\scalebar^2\vargauss^2)^2}\cdot \E{\indicator\{\scalebar^2 \geq \tau\}}\Big)^{\nicefrac{1}{2}}\nonumber\\ 
    &\overset{\stepfour}{\leq} \Big( \E{|\scalebar^2|^4}\cdot \E{|\vargauss^2|^4} \Big)^{\nicefrac{1}{4}} \Big(\P{\scalebar^2 \geq \tau}\Big)^{\nicefrac{1}{2}},\label{eq:prop2_term2_op_norm_bound}
\end{align}
where where step~\stepone is true because $\scalebar^2 \ge \scaletilde^2$ from to~\eqref{tp2}, step~\steptwo is true due to~\eqref{tp3}, and steps~\stepthree and~\stepfour follow from Cauchy--Schwarz inequality. We proceed by bounding each of the terms in~\eqref{eq:prop2_term2_op_norm_bound} separately. First, we can upper bound the fourth moment $\E{\vert \scalebar^2 \vert^4}$ by the quantity $\fourthmoment$ from Equation~\eqref{eq:fourth_moment_recall_prop2}. Second, $\vargauss^2$ is a sub-exponential random variable. By Lemma~\ref{lemma:subexp_moments} in Appendix~\ref{app:prelim:sub_exp}, we have that $\E{\vert \vargauss^2 \vert^4}^{\nicefrac{1}{4}} \leq c$ for some constant $c$. It remains to bound the term $\Big(\P{\scalebar^2 \geq \tau}\Big)^{\nicefrac{1}{2}}$. We have
\begin{align}
    \P{\scalebar^2 \geq \tau} &\overset{\stepone}{\leq} \frac{\Expect\;\abs{\scalebar^2}^2}{\trunc^2}\nonumber\\ 
    &\overset{\steptwo}{\leq} \frac{\left(\Expect\; \abs{\scalebar^2}^4\right)^{\nicefrac{1}{2}}}{\trunc^2}\nonumber\\
    &\overset{\stepthree}{\leq} \frac{\fourthmoment^{\nicefrac{1}{2}}}{\trunc^2},\nonumber
\end{align}
where step~\stepone follows from Markov's inequality, step~\steptwo follows from Cauchy--Schwarz inequality, and step~\stepthree follows from the fourth moment bound on the averaged scaling $\scalebar^2$. Putting everything together back to~\eqref{eq:prop2_term2_op_norm_bound}, we have 
\begin{align}
\abs*{\vv^\top \left( \E{\mAbar} - \E{\mAtilde} \right)\vv }\lesssim \frac{\fourthmoment^{\nicefrac{1}{2}}}{\trunc}\nonumber
\end{align}
for any vector $\vv \in \sphere^{d-1}$. Therefore, 
\begin{align}
    \opnorm{\E{\mAtilde} - \E{\mAbar}} \lesssim \frac{\fourthmoment^{\nicefrac{1}{2}}}{\trunc},\nonumber
\end{align}
as desired.

\subsubsection{Proof of Lemma~\ref{lemma:op_bound_3}}\label{sec:op_bound_3}
Substituting in the definitions $\mAbar = \scalebar^2\opa{}$ and $\mAtilde = \scaletilde^2\opa{}$, we have \begin{align}
    \ip{\mAbar}{\Sigstar}\mAbar -\ip{\mAtilde}{\Sigstar}\mAtilde= \left(\scalebar^4 - \scaletilde^4\right)\left(\quadSig{}\right) \opa{}.\nonumber
\end{align}
Therefore, our goal is to bound the operator norm
\begin{align}
    \opnorm{\left(\scalebar^4 - \scaletilde^4\right)\left(\quadSig{}\right) \opa{}} = \sup_{\vv\in \sphere^{\dimension-1}} \abs*{\vv^T \left(\scalebar^4 - \scaletilde^4\right)\left(\quadSig{}\right) \opa{}\vv}.\nonumber
\end{align}

Similar to the proof of Lemma~\ref{lemma:op_bound_2}, we fix any vector $\vv\in \sphere^{\dimension-1}$. 
Again, note that $\vv^\top\va \sim \normal(0,1)$ and denote $\vargauss \sim \normal(0,1)$. We have
\begin{align}
    \abs*{\vv^\top\E{\left(\scalebar^4 - \scaletilde^4\right)\left(\quadSig{}\right) \opa{}}\vv}
    &\stackrel{\stepone}{=} \E{\left(\scalebar^4 - \scaletilde^4\right)\left(\quadSig{}\right) \vargauss^2}\nonumber\\
    &= \E{\left(\scalebar^2 + \scaletilde^2\right)\left(\scalebar^2 - \scaletilde^2\right)\left(\quadSig{}\right) \vargauss^2}\nonumber\\
    &\overset{\steptwo}{\leq} \E{2\scalebar^2\left(\scalebar^2 - \scaletilde^2\right)\left(\quadSig{}\right) \vargauss^2}\nonumber\\
    &\overset{\stepthree}{=} 2\E{(\boundary + \noisebar)\left(\scalebar^2 - \scaletilde^2\right) \vargauss^2 }\nonumber\\
    &\overset{\stepfour}{\leq} 2(\boundary + \noiseup)\E{\left(\scalebar^2 - \scaletilde^2\right) \vargauss^2 \vone\{\scale^2 \geq \tau\}}\nonumber
\end{align}
where steps~\stepone and~\steptwo are true because $\scalebar^2 \ge \scaletilde^2$ from~\eqref{tp2}, step~\stepthree follows from the definition $\scalebar^2 = \frac{y + \noisebar}{\quadSig{}}$, and step~\stepfour follows from~\eqref{tp3} and the definition of $\noiseup$ as the upper bound on the noise $\noise$.

The rest of the proof follows the exact steps of the proof of Lemma~\ref{lemma:op_bound_2} in Section~\ref{sec:op_bound_2}. Therefore, we have the bound
\begin{align}
\opnorm{\E{\left(\scalebar^4 - \scaletilde^4\right)\left(\quadSig{}\right) \opa{}}} \lesssim \frac{\boundaryup \fourthmoment^{\nicefrac{1}{2}}}{\tau},\nonumber
\end{align}
as desired.

\subsubsection{Proof of Lemma~\ref{lemma:op_bound_4}}\label{sec:op_bound_4}
The proof follows the steps as in the proof of Lemma~\ref{lemma:op_bound_1}, and we describe the difference of the two proofs. We again apply Bernstein's inequality. 

\paragraph{Proving a Bernstein condition.} We prove a Bernstein condition with $\constbernone = \const_1 (\boundary + \noiseup)^2$ and $\constberntwo = \const_2 (\boundary + \noiseup) \tau$. Namely, for every integer $\power \ge 2$, we have (cf.~\eqref{eq:term_1_bern_condition} in \Cref{lemma:op_bound_1})
\begin{align}
    \E{\left|\vv^\top\ip{\mAtilde}{\Sigstar}\mAtilde\vv\right|^p} & \le \frac{\power!}{2} \constbernone \constberntwo^{\power-2}.\label{eq:term_4_bern_condition}
\end{align}
To show~\eqref{eq:term_4_bern_condition}, we plug in $\mAtilde = \scaletilde^2 \opa{}$ and have
\begin{align}
    \Expect \;\left|\vv^\top\ip{\mAtilde}{\Sigstar}\mAtilde\vv\right|^p &= 
    \Expect\; \left(\scaletilde^2 \va^\top\Sigstar\va\right)^\power\cdot \abs*{\vv^\top\mAtilde\vv}^\power \nonumber\\
    & \overset{\stepone}{\leq} \Expect\; \left(\scalebar^2 \va^\top\Sigstar\va\right)^\power\cdot \abs*{\vv^\top\mAtilde\vv}^\power\nonumber\\
    & \overset{\steptwo}{=} \Expect\; \left(\boundary + \noisebar\right)^\power\cdot \abs*{\vv^\top\mAtilde\vv}^\power\nonumber\\
    & \overset{\stepthree}{\leq} (\boundary + \noiseup)^\power \cdot \Expect\; \abs*{\vv^\top \mAtilde\vv}^p,\label{eq:term_4_bern_condition_expression}
\end{align}
where step~\stepone follows from~\eqref{tp2}, step~\steptwo follows from the definition $\scalebar^2 = \frac{y + \noisebar}{\quadSig{}}$, and step~\stepthree follows from the definition of $\noiseup$ as the upper bound on the noise $\noise$. 
Substituting in~\eqref{eq:term_1_bern_condition} from Lemma~\ref{lemma:op_bound_1} to bound the term $\Expect\;\abs*{\vv^\top \mAtilde\vv}^\power$ in~\eqref{eq:term_4_bern_condition_expression} completes the proof of the Bernstein condition~\eqref{eq:term_4_bern_condition}.

\paragraph{Applying Bernstein's inequality.}
The rest of the proof follows in the same manner as the proof of Lemma~\ref{lemma:op_bound_1} in \Cref{sec:op_bound_1}, with an additional factor of $(\boundary + \noiseup)$. We have
\begin{align}
    \opnorm{\E{\ip{\mAtilde}{\Sigstar}\mAtilde} - \frac1{n}\sum\limits_{i=1}^n\ip{\mAtilde_i}{\Sigstar}\mAtilde_i} \lesssim \boundaryup\left(\sqrt{\frac{\fourthmoment^{\nicefrac{1}{2}}t}{n}} + \frac{\tau t}{n}\right)\nonumber
\end{align}
with probability at least $1 - 2 \cdot 9^d \cdot \exp\left(-t\right)$, as desired.

\subsubsection{Proof of Lemma~\ref{lemma:op_bound_5}}\label{sec:op_bound_5}
Recall that by definition $\mAbar =\scalebar^2 \opa{} = \frac{\boundary + \noisebar}{\va^\top \Sigstar \va}\opa{}$. We have 
\begin{align}
\opnorm{\E{\noisebar\mAbar}} &= \opnorm{\E{\noisebar(\boundary+\noisebar)\frac{\opa{}}{\quadSig{}}}} \nonumber\\
&= \opnorm{\E{\noisebar(\boundary+\noisebar)}\cdot \E{\frac{\opa{}}{\quadSig{}}}} \nonumber\\
&= \frac{\sigma_\eta^2}{m}\cdot \opnorm{\E{\frac{\opa{}}{\quadSig{}}}}.\label{eq:term_5_decompose}
\end{align}
To bound the operator norm term in~\eqref{eq:term_5_decompose}, we apply \Cref{lemma:quad_form_moments}\ref{part:quad_form_ratio} in Appendix~\ref{app:prelim:bao_kan}. For any matrix $\varmtx$, we have
\begin{align}
    \Expect\left[\frac{\va^\top \varmtx\va}{\va^\top \Sigstar \va}\right] \lessorder \frac{1}{\singularvalmin\rank}\normnuc{\varmtx}.\label{eq:lem_op_bound_5_apply_ratio}
\end{align}
Note that $\frac{\opa{}}{\quadSig{}}$ is symmetric positive semidefinite, so we have
\begin{align}
    \opnorm{\E{\frac{\opa{}}{\quadSig{}}}} & = \sup_{\vv\in \sphere^{\dimension -1}} \abs*{\vv^\top \E{\frac{\opa{}}{\quadSig{}}} \vv } \nonumber\\
    & = \sup_{\vv\in \sphere^{\dimension -1}} \E{\frac{\va^\top (\vv \vv^\top) \va}{\quadSig{}}} \nonumber \\
    & \stackrel{\stepone}{\lessorder} \frac{1}{\singularvalmin \rank}\sup_{\vv\in \sphere^{\dimension -1}} \normnuc{\vv\vv^\top}\nonumber\\
    & \stackrel{\steptwo}{=} \frac{1}{\singularvalmin \rank}\label{eq:lem_op_bound_5_bound_term},
\end{align}
where step~\stepone is true by substituting in~\eqref{eq:lem_op_bound_5_apply_ratio} with $\varmtx = \vv\vv^T$, and step~\steptwo is true because $\vv$ is unit norm, and hence $\nucnorm{\vv\vv^\top} = 1$. 
Substituting~\eqref{eq:lem_op_bound_5_bound_term} back to~\eqref{eq:term_5_decompose}, we have
\begin{align}
    \opnorm{\E{\noisebar \mAbar}} \lesssim \frac{1}{\sigma_r r} \cdot \frac{\noisevar}{m},\nonumber
\end{align}
as desired.

\subsection{Proof of Proposition~\ref{prop:rsc}}\label{sec:proof-rsc}
We analyze the term $\frac{1}{n} \sum\limits_{i=1}^n \ip{\mAtilde_i}{\mU}^2$ from~\eqref{eq:prop_rsc_condition}. Recall from the definition of $\mAtilde$ that for any $i = 1, \ldots, n$,
\begin{align}
    \mAtilde_i &= \scaletilde_i^2 \opa{i}
    = \left( \frac{\boundary + \noisebar_i}{\quadSig{i}} \wedge \trunc \right) \opa{i},\nonumber
\end{align}
so we have
\begin{align}
    \ip{\mAtilde_i}{\mU}^2 = \left( \frac{\boundary + \noisebar_i}{\quadSig{i}} \wedge \trunc \right)^2 \left( \quadmU{i} \right)^2.\label{eq:matrix_product_squared_expression}
\end{align}
From~\eqref{eq:matrix_product_squared_expression}, we have that for any matrix $\mU$, the term $\sum\limits_{i=1}^n \ip{\mAtilde_i}{\mU}^2$ is nondecreasing in $\trunc$ when $\trunc > 0$. Defining a random matrix
\begin{align}\label{eq:trunc_Atilde}
    \mAtrunc \defn \left( \frac{\boundary + \noisebar}{\quadSig{}} \wedge \truncprime \right) \opa{},
\end{align}
for any $\truncprime\in (0,  \trunc]$, we have
\begin{align}\label{eq:lem_small_ball_monotonicity}
    \frac{1}{n} \sum\limits_{i=1}^n \ip{\mAtilde_i}{\mU}^2 \geq \frac{1}{n} \sum\limits_{i=1}^n \ip{\mAtrunc_i}{\mU}^2,
\end{align}
where for every $i = 1, \ldots, n$, matrix $\mAtrunc_i$ is formed with the same realizations of random quantities $\va_i$ and $\noisebar_i$ as $\mAtilde_i$. The two matrices only differ in choice of truncation threshold: $\truncprime$ instead of $\trunc$. As a result, for the rest of the proof, we lower bound $\frac{1}{n} \sum\limits_{i=1}^n \ip{\mAtrunc_i}{\mU}^2$ for an appropriate choice of $\truncprime$ to be specified later. To proceed, we use a small-ball argument \cite{mendelson2015learning, tropp2015convex} based on the following lemma.

\begin{lemma}[{\cite[Proposition 5.1]{tropp2015convex}}, adapted to our notation]
    \label{lemma:small-ball}
    Let $\mX_1, \dots, \mX_n \in \bbR^{d \times d}$ be i.i.d.\ copies of a random matrix $\mX \in \bbR^{d \times d}$.
    Let $E \subset \bbR^{d \times d}$ be a subset of matrices.
    Let $\xi > 0$ and $Q > 0$ be real values such that for every matrix $\mU \in E$, the marginal tail condition holds:
    \begin{align}   
        \P{\abs{\ip{\mX}{\mU}} \geq 2\xi} \geq Q.\label{eq:lem_tropp_condition}
    \end{align}
    Define the Rademacher width as
    \begin{align}        
        W \defn \E{ \sup_{\mU \in E}~ \frac{1}{n} \sum_{i=1}^n \varepsilon_i \ip{\mX_i}{\mU} },\nonumber
    \end{align}
    where $\varepsilon_1, \dots, \varepsilon_n$ are i.i.d.\ Rademacher random variables independent of $\{\mX_i\}_{i\in [\numitems]}$.
    Then for any $t > 0$, we have
    \[
        \inf_{\mU \in E}~\left( \frac{1}{n} \sum_{i=1}^n \ip{\mX_i}{\mU}^2 \right)^{1/2} \geq \xi ( Q - t ) - 2 W.
    \]
    with probability at least $1 - \exp\left(-\frac{nt^2}{2}\right)$.
\end{lemma}
Recall the error set $\calE$ defined in Equation~\eqref{eq:error_set}. Because the claim~\eqref{eq:prop_rsc_condition} is invariant to scaling, it suffices to prove it for $\fronorm{\mU} = 1$. Correspondingly, we define the set $E$ as
\begin{align}
    E &= \calE \cap \{\mU \in \bbR^{d \times d} : \fronorm{\mU} = 1 \} \nonumber\\
    &= \{ \mU \in \bbS^{d \times d} : \fronorm{\mU} = 1, \nucnorm{\mU} \leq 4\sqrt{2r} \}. \label{eq:error_set_fronorm}
\end{align}

We invoke Lemma~\ref{lemma:small-ball} with set $E$ defined above, $\mX_i = \mAtrunc_i$, $\xi = \frac{c_1}{2}\left(\frac{\noisemedian}{\tr{\Sigstar}} \wedge \truncprime \right)$, and $Q = c_2$, where $\noisemedian$ is the median of $\noisebar$ and $c_1$ and $c_2$, are constants to be specified later. The rest of the proof is comprised of two steps. We first verify that our choices for $\xi$ and $Q$ are valid for establishing the marginal tail condition~\eqref{eq:lem_tropp_condition}. We then bound the Rademacher width $W$ above. The following lemma verifies our choices for $\xi$ and $Q$.

\begin{lemma}\label{lemma:small-ball-Q}
    Consider any $\truncprime \in (0, \trunc]$. There exist absolute constants $c_1, c_2 > 0$ such that for every $\mU \in E$, we have
    \begin{align}
        \P{\left|\ip{\mAtrunc}{\mU} \right| \geq c_1\left(\frac{\noisemedian}{\tr{\Sigstar}}\wedge \truncprime\right) } \geq c_2.\nonumber
    \end{align}
\end{lemma}

The proof of Lemma~\ref{lemma:small-ball-Q} is presented in Section~\ref{sec:small-ball-Q}. We now turn to the second step of the proof, which is bounding the Rademacher width $W$. The next lemma characterizes this width.
\begin{lemma}\label{lemma:small-ball-rad-width}
    Consider any $\truncprime\in (0, \trunc]$. Let $\mAtrunc_1, \ldots, \mAtrunc_n \in \bbR^{d \times d}$ be i.i.d. copies of the random matrix $\mAtrunc \in \bbR^{d \times d}$ defined in Equation~\eqref{eq:trunc_Atilde}. Let $E$ be the set defined in Equation~\eqref{eq:error_set_fronorm}. Then, there exists some absolute constants $c_1$ and $c_2$ such that if $n \geq c_1 d$, then we have
    \begin{align}
        \E{ \sup_{\mU \in E}~ \frac{1}{n} \sum_{i=1}^n \varepsilon_i \ip{\mAtrunc_i}{\mU} } \leq c_2\truncprime\sqrt{\frac{rd}{n}}.\nonumber
    \end{align}
\end{lemma}

The proof of Lemma~\ref{lemma:small-ball-rad-width} is presented in Section~\ref{sec:small-ball-rad-width}. Lemma~\ref{lemma:small-ball-Q} establishes the marginal tail condition for Lemma~\ref{lemma:small-ball}, and Lemma~\ref{lemma:small-ball-rad-width} upper bounds the Rademacher width. We now invoke Lemma~\ref{lemma:small-ball} and substitute in the upper bound for the Rademacher width $W$. For some constant $c_4$, if $n \geq c_4 d$, we have that with probability at least $1 - \exp\left(-\frac{nt^2}{2}\right)$,
\begin{align}
    \inf_{\mU \in E} \left(\frac{1}{n} \sum\limits_{i=1}^n \ip{\mAtilde_i}{\mU}^2\right)^{\nicefrac{1}{2}} & \stackrel{\stepone}{\geq} \inf_{\mU \in E} \left(\frac{1}{n} \sum\limits_{i=1}^n \ip{\mAtrunc_i}{\mU}^2\right)^{\nicefrac{1}{2}} \nonumber\\
    &\geq \frac{c_1}{2}\left(\frac{\noisemedian}{\tr{\Sigstar}}\wedge \truncprime\right) \left(c_2 - t \right) - c_3\truncprime\sqrt{\frac{rd}{n}},\nonumber
\end{align}
where step~\stepone is true due to the monotonicity property~\eqref{eq:lem_small_ball_monotonicity}.
We set $\truncprime = \frac{\noisemedian}{\tr{\Sigstar}}$, where recall that $\noisemedian$ is the median of the random quantity $\boundary + \noisebar$. By the assumption $\trunc \geq \frac{\noisemedian}{\tr{\Sigstar}}$, this choice of $\truncprime$ satisfies $\truncprime \leq \trunc$.
Setting $t = \frac{c_2}{2}$, we have that with probability at least $1 - \exp\left(-\frac{c_2^2 n}{8}\right)$,
\begin{align}
    \inf_{\mU \in E} \frac{1}{n} \left(\sum\limits_{i=1}^n \ip{\mAtrunc_i}{\mU}^2\right)^{\nicefrac{1}{2}} &\geq \frac{\const_1\const_2}{4} \frac{\noisemedian}{\tr{\Sigstar}} - \const_3\frac{\noisemedian}{\tr{\Sigstar}}\sqrt{\frac{rd}{n}}.\nonumber
\end{align}
Recall from the definition of $E$~\eqref{eq:error_set_fronorm} that $\fronorm{\mU} = 1$. As a result, if $n \geq \max\left\{\left(\frac{4c_3}{c_1c_2}\right)^2, c_4\right\}rd$, we have
\begin{align}
    \inf_{\mU \in \calE} \frac{1}{n} \sum\limits_{i=1}^n \ip{\mAtilde_i}{\mU}^2 \geq \left(\frac{c_1c_2}{4}\frac{\noisemedian}{\tr{\Sigstar}}\right)^2 \fronorm{\mU}^2\nonumber
\end{align}
with probability at least $1 - \exp\left(-\frac{c_2^2 n}{8}\right)$. We conclude by setting $\kappa_\calL = \left(\frac{c_1c_2}{4}\right)^2$, $c = \frac{c^2_2}{8}$, and $C = \max\left\{\left(\frac{4c_3}{c_1c_2}\right)^2, c_4\right\}$ in \Cref{prop:rsc}.

\subsubsection{Proof of Lemma~\ref{lemma:small-ball-Q}}\label{sec:small-ball-Q}

We fix any $\mU \in E$. Recall that $\noisemedian$ denotes the median of $\boundary + \noisebar$. Let $\calG$ be the event that $\boundary + \noisebar \geq \noisemedian$, which occurs with probability $\frac12$. For any $\xi > 0$, because the averaged noise $\noisebar$ and sensing vector $\va$ are independent, we have
\begin{align}
     \P{\left|\ip{\mAtrunc}{\mU} \right| \geq \xi} & \stackrel{\stepone}{=} \P{\left(\frac{\boundary + \noisebar}{\quadSig{}} \wedge \truncprime \right)\cdot \left| \ip{\opa{}}{\mU} \right| \geq \xi} \nonumber\\
     &= \P{\left(\frac{\boundary + \noisebar}{\quadSig{}} \wedge \truncprime \right) \cdot \left| \ip{\opa{}}{\mU} \right| \geq \xi \bigg\vert\; \calG} \cdot \P{\calG}\nonumber\\
     &= \frac12 \P{\left(\frac{\boundary + \noisebar}{\quadSig{}} \wedge \truncprime \right) \cdot \left| \ip{\opa{}}{\mU} \right| \geq \xi \bigg\vert\; \calG} \nonumber\\
     & \stackrel{\steptwo}{\geq} \frac12 \P{ \left(\frac{\noisemedian}{\quadSig{}} \wedge \truncprime \right) \cdot \left| \ip{\opa{}}{\mU} \right| \geq \xi},\label{eq:lem_tropp_expression}
\end{align}
where step~\stepone is true by plugging in the definition of $\mAtrunc$, and step~\steptwo is true by the definition of the event $\calG$.
We proceed by bounding the terms in~\eqref{eq:lem_tropp_expression} separately. 

\paragraph{Lower bound on $\abs*{\ip{\opa{}}{\mU}}$.}
We use the approach from \cite[Section 4.1]{kueng2017low}. By Paley-Zygmund inequality,
\begin{align}\label{eq:paley_zygmund}
    \P{\left| \ip{\opa{}}{\mU} \right|^2 \geq \frac{1}{2}\E{\left| \ip{\opa{}}{\mU} \right|^2}  } \geq \frac14 \frac{\left(\E{\left| \ip{\opa{}}{\mU} \right|^2}\right)^2}{\E{\left| \ip{\opa{}}{\mU} \right|^4}}
\end{align}
We now analyze the terms in~\eqref{eq:paley_zygmund}. As noted in \cite[Section 4.1]{kueng2017low}, there exists some constant $c_1 > 0$ such that for any matrix $\mU$ with $\normfro{\mU} = 1$,
\begin{align}\label{eq:paley_zygmund_bounds}
    \E{\abs*{\ip{\opa{}}{\mU}}^2} \geq 1 \quad \text{ and } \quad \E{\abs*{\ip{\opa{}}{\mU}}^4} \leq  c_1 \left( \E{\abs*{\ip{\opa{}}{\mU}}^2} \right)^2.
\end{align}
Note that by the definition of the set $E$, every matrix $\mU \in E$ satisfies $\normfro{\mU} = 1$. Utilizing inequalities~\eqref{eq:paley_zygmund} and~\eqref{eq:paley_zygmund_bounds}, there exists positive constant $c_2 > 0$ such that
\begin{align}
    \P{\left| \ip{\opa{}}{\mU} \right| \geq \frac{1}{2}} \geq c_2.\label{eq:lem_tropp_lb}
\end{align}
\paragraph{Upper bound on $\quadSig{}$.}
By Hanson-Wright inequality~\cite[Theorem 1.1]{rudelson2013hanson}, there exist some positive absolute constants $c_3$ and $c_4$ such that for any $t > 0$, we have 
\begin{align}
    \Prob\left(\quadSig{} \leq c_3\left( \tr{\Sigstar} + \fronorm{\Sigstar}\sqrt{t} + \opnorm{\Sigstar}t \right)\right) \ge 1 - 2\exp\left(-c_4t\right). \nonumber
\end{align}
We set $t = -\frac{1}{c_4}\log(\frac{c_2}{4})$ so that $2\exp\left(-c_4t\right) = \frac{c_2}{2}$. Since $\Sigstar$ is symmetric positive semidefinite, we have
\begin{align*}
    & \fronorm{\Sigstar} \leq \tr{\Sigstar}\\
    \text{and } & \opnorm{\Sigstar} \leq \tr{\Sigstar}
\end{align*}
As a result, we have that there exists some constant $c_5 > 0$ such that
\begin{align}
    \Prob\Big(\quadSig{} \leq c_5\tr{\Sigstar}\Big)\ge 1 - \frac{c_2}{2}.\label{eq:lem_tropp_ub}
\end{align}

\paragraph{Substituting the two bounds back to~\eqref{eq:lem_tropp_expression}.}
By a union bound of~\eqref{eq:lem_tropp_lb} and~\eqref{eq:lem_tropp_ub}, we have 
\begin{align}
    &\P{ \left(\frac{\noisemedian}{\quadSig{}} \wedge \truncprime \right)\cdot \left| \ip{\opa{}}{\mU} \right|
    \geq \frac{1}{2} \left(\frac{\noisemedian}{c_5\tr{\Sigstar}} \wedge \truncprime\right)} \nonumber\\ 
    &\qquad \geq \P{\frac{\noisemedian}{\quadSig{}} \wedge \truncprime \geq \frac{\noisemedian}{c_5\tr{\Sigstar}} \wedge \truncprime} + \P{\left| \ip{\opa{}}{\mU} \right| \geq \frac{1}{2}} - 1 \nonumber\\
    & \qquad \ge \P{\frac{\noisemedian}{\quadSig{}}  \geq \frac{\noisemedian}{c_5\tr{\Sigstar}} } + \P{\left| \ip{\opa{}}{\mU} \right| \geq \frac{1}{2}} - 1 \geq \frac{c_2}{2}\label{eq:lem_tropp_combine}
\end{align}
Combining~\eqref{eq:lem_tropp_combine} and~\eqref{eq:lem_tropp_expression}, and redefining constant $c_2$ appropriately, we have
\begin{align}
    \P{\left|\ip{\mAtrunc}{\mU} \right| \geq \frac{1}{2}\left(\frac{\noisemedian}{\tr{\Sigstar}} \wedge \truncprime\right) } \geq c_2,\nonumber
\end{align}
as desired.

\subsubsection{Proof of Lemma~\ref{lemma:small-ball-rad-width}}\label{sec:small-ball-rad-width}
We begin by noting that for any matrix $\mU \in E$, 
\begin{align}
    \E{ \sup_{\mU \in E}~ \frac{1}{n} \sum_{i=1}^n \varepsilon_i \ip{\mAtrunc_i}{\mU} } &\overset{\stepone}{\leq} \E{ \sup_{\mU \in E}~ \opnorm{\frac{1}{n} \sum_{i=1}^n \varepsilon_i \mAtrunc_i}\cdot \nucnorm{\mU} } \nonumber\\
    &\overset{\steptwo}{\leq} 4\sqrt{2r}\cdot  \Expect\; \opnorm{\frac{1}{n} \sum_{i=1}^n \varepsilon_i \mAtrunc_i},\label{eq:lem_rademacher_op_expression}
\end{align}
where step~\stepone follows from H\"{o}lder's inequality, and step~\steptwo follows from the fact that $\normnuc{\mU} \le 4\sqrt{2\rank}$ from the definition of the set $E$. It remains to bound the expectation of the operator norm in~\eqref{eq:lem_rademacher_op_expression}. We follow the standard covering arguments in \cite[Section 5.4.1]{vershynin2010introduction}, \cite[Section 8.6]{tropp2015convex}, \cite[Section 4.1]{kueng2017low}, with a slight modification to accommodate the bounded term $\left(\frac{\boundary + \noisebar_i}{\quadSig{i}} \wedge \truncprime \right)$ that appears in each of the matrices $\mAtrunc_i$. As a result, there exist universal constants $c_1$ and $c_2$ such that if $n$ satisfies $n \geq c_1 d$, then we have
\begin{align}
    \E{ \opnorm{\frac{1}{n} \sum_{i=1}^n \varepsilon_i \mAtrunc_i}} \leq c_2\truncprime\sqrt{\frac{d}{n}}. \nonumber
\end{align}
We conclude by re-defining $c_2$ appropriately. 

\section{Proof of Corollary~\ref{cor:upper_bounds_rate}}\label{sec:cor1_proof}
We proceed by considering two cases. For each case, the proof consists of two steps. We first verify that the choices of the averaging parameter $m$ and truncation threshold $\trunc$, 
\begin{align}
    m = \Bigg\lceil \left[\left(\frac{\noisevar}{(\boundaryup)^2}\right)^2\frac{N}{d}\right]^{\nicefrac13}\Bigg\rceil \quad \text{ and } \quad \trunc = \frac{\boundaryup}{\sigma_r r}\sqrt{\frac{N}{md}}, \label{eq:cor_vals_m_tau} 
\end{align}
satisfy the assumptions of Theorem~\ref{theorem:upper_bounds}, namely $n \geq C_2 rd$ and $\trunc \geq \frac{\noisemedian}{\tr{\Sigstar}}$. We then invoke Theorem~\ref{theorem:upper_bounds}.

\subsection{Case 1: high-noise regime} 
In this case, we have $\frac{\noisevar}{(\boundaryup)^2} > \sqrt{\frac{d}{N}}$, which means by setting $m$ according to Equation~\eqref{eq:cor_vals_m_tau}, we have $m \geq 2$. As a result, the bound
\begin{align}\label{eq:high_noise_ceil_bound}
    \Bigg\lceil \left[\left(\frac{\noisevar}{(\boundaryup)^2}\right)^2\frac{N}{d}\right]^{\nicefrac13}\Bigg\rceil \leq 2 \left[\left(\frac{\noisevar}{(\boundaryup)^2}\right)^2\frac{N}{d}\right]^{\nicefrac13}
\end{align}
holds in the high-noise regime.
\paragraph{Verifying the condition on $n$.}
Recall that $n = \frac{N}{m}$. We have
\begin{align}
    n = \frac{N}{m} &\stackrel{\stepone}{\geq} \frac{N}{2} \left( \left(\frac{\noisevar}{(\boundaryup)^2}\right)^2\frac{N}{d} \right)^{-1/3} \nonumber\\
    &= \frac{1}{2}\left( N^2 d \left(\frac{(\boundaryup)^2}{\noisevar}\right)^2 \right)^{1/3} \nonumber\\
    & \stackrel{\steptwo}{\geq} \left( C_2^3 \left(\frac{(\boundaryup)^2}{\noisevar}\right)^2 \left(\frac{\noisevar}{(\boundaryup)^2}\right)^2 r^3 d^3   \right)^{1/3} \nonumber\\
    &= C_2 rd,\nonumber
\end{align}
where step~\stepone is true by plugging in the choice of $m$ from~\eqref{eq:cor_vals_m_tau} and applying the bound~\eqref{eq:high_noise_ceil_bound}, and step~\steptwo is true by substituting in the assumption $N \geq 2C_2^{\nicefrac{3}{2}} \left(\frac{\noisevar}{(\boundaryup)^2}\right)^2 r^{\nicefrac{3}{2}}d$. Thus the condition $n \geq C_2 rd$ of \Cref{theorem:upper_bounds} is satisfied.

\paragraph{Verifying the condition on $\trunc$.} For the term $\sqrt{\frac{N}{dm}}$ in the expression of $\trunc$ in~\eqref{eq:cor_vals_m_tau},
note that, by the previous point, $\frac{N}{m} = n \gtrsim rd$ (with a constant that is greater than 1). Thus $\sqrt{\frac{N}{dm}} \geq \sqrt{r} > 1$.
Therefore, to verify the condition $\trunc \geq \frac{\noisemedian}{\tr{\Sigstar}}$, it suffices to verify that \begin{align}
    \frac{\boundaryup}{\sigma_r r} \geq \frac{\noisemedian}{\tr{\Sigstar}}.\label{eq:cor_tau_condition}
\end{align}
By definition, we have $\boundaryup \geq \noisemedian$. Furthermore, since $\Sigstar$ is symmetric positive semidefinite, its eigenvalues are all non-negative and are identical to its singular values, and hence $\tr{\Sigstar} \ge \sigma_r r$, verifying the condition~\eqref{eq:cor_tau_condition}.

\paragraph{Invoking Theorem~\ref{theorem:upper_bounds}.} By setting $\lambda_n$ to its lower bound in~\eqref{eq:reg_n_conditions} and substituting in $n = \frac{N}{m}$ and our choice of $\trunc$ from~\eqref{eq:cor_vals_m_tau}, we have
\begin{align}
    \lambda_n = C_1\left(3\frac{(\boundaryup)^2}{\sigma_r r}\sqrt{\frac{md}{N}} + \frac{\noisevar}{m}\right) \label{eq:cor_val_tau_prelim}
\end{align}
Substituting this expression of $\lambda_n$ to the error bound~\eqref{eq:error_bound}, then substituting in our choice of $m$ from~\eqref{eq:cor_vals_m_tau} to~\eqref{eq:cor_val_tau_prelim} and defining $C^\prime = 3C \cdot C_1$, we have
\begin{align}
    \fronorm{\Sighat - \Sigstar} \leq C^\prime \left(\frac{\tr{\Sigstar}^2}{\sigma_r r}\right) \frac{(\boundaryup)^{\nicefrac{4}{3}} (\noisevar)^{\nicefrac{1}{3}}}{\noisemedian^2} \sqrt{r}\left(\frac{d}{N}\right)^{\nicefrac{1}{3}}. \nonumber
\end{align}
Using the fact that $\tr{\Sigstar} \leq \sigma_1 r$, we have
\begin{align}
    \fronorm{\Sighat - \Sigstar} \leq C^\prime\ \frac{\sigma_1^2}{\sigma_r} \frac{(\boundaryup)^{\nicefrac43} (\noisevar)^{\nicefrac13}}{\noisemedian^2}\ r^{\nicefrac{3}{2}}\left(\frac{d}{N} \right)^{\nicefrac{1}{3}} \nonumber
\end{align}
as desired.

\subsection{Case 2: low-noise regime}
In this case, we have $\frac{\noisevar}{(\boundaryup)^2} \leq \sqrt{\frac{d}{N}}$, which means by setting $m$ according to Equation~\eqref{eq:cor_vals_m_tau}, we have $m = 1$. As a result, no averaging occurs.

\paragraph{Verifying the condition on $n$.} Because $m = 1$ in this case, we have that $n = N$. By assumption, we have that $N \geq C_2rd$, satisfying the condition $n \geq C_2rd$ in Theorem~\ref{theorem:upper_bounds}. 

\paragraph{Verifying the condition on $\trunc$.} By the same analysis as in Case 1, we have that the condition $\trunc \geq \frac{\noisemedian}{tr{\Sigstar}}$ in Theorem~\ref{theorem:upper_bounds}. 

\paragraph{Invoking Theorem~\ref{theorem:upper_bounds}.} By setting $\lambda_n$ to its lower bound in ~\eqref{eq:reg_n_conditions}, substituting in our choice of $\trunc$ from~\eqref{eq:cor_vals_m_tau} and noting $m = 1$, we have
\begin{align}\label{eq:cor_lambda_lb_trunc}
    \lambda_n = C_1\left(3\frac{(\boundaryup)^2}{\sigma_r r}\sqrt{\frac{d}{n}} + \frac{1}{\sigma_r r}\noisevar\right).
\end{align}

We define $C^\prime = 3C \cdot C_1$ and note that $n = N$ under Case 2. Substituting this expression of $\lambda_n$ in~\eqref{eq:cor_lambda_lb_trunc} to the error bound~\eqref{eq:error_bound}, then using the fact that under Case 2, the bound $\noisevar \leq (\boundaryup)^2\sqrt{\frac{d}{N}}$ holds, we have
\begin{align*}
    \fronorm{\Sighat - \Sigstar} \leq C^\prime \left(\frac{\tr{\Sigstar}^2}{\sigma_r r}\right) \left(\frac{\boundaryup}{\noisemedian}\right)^2\sqrt{\frac{rd}{N}}.
\end{align*}
Using the fact that $\tr{\Sigstar} \leq \sigma_1 r$, we have
\begin{align*}
    \fronorm{\Sighat - \Sigstar} \leq C^\prime\ \frac{\sigma_1^2}{\sigma_r} \left(\frac{\boundaryup}{\noisemedian}\right)^2\sqrt{\frac{r^3d}{N}},
\end{align*}
as desired.

\subsection*{Acknowledgments}
AX was supported by National Science Foundation grant CCF-2107455. JW was supported by the Ronald J. and Carol T. Beerman President’s Postdoctoral Fellowship and the Algorithms and Randomness Center Postdoctoral Fellowship at Georgia Tech. MD was supported by National Science Foundation grants CCF-2107455, IIS-2212182, and DMS-2134037. AP was supported in part by the National Science Foundation grants CCF-2107455 and DMS-2210734, and by research awards/gifts from Adobe, Amazon and Mathworks.

\bibliographystyle{plain}
\bibliography{refs}

\appendix

\section{Simulation details}\label{app:sims}

In this section, we provide details for the simulation results presented in Figure~\ref{fig:noiseless_sims}. For our experiments, we adopt a normalized version of the setup of~\cite{chen2015exact} and form the metric $\Sigstar$ by $\Sigstar = \mL\mL^\top / \fronorm{\mL\mL^\top}$, where $\mL$ is a $50 \times 10$ matrix with i.i.d. Gaussian entries. We sweep the number of query responses $N$, estimate the metric with $\Sighat$, and report the normalized estimation error $\fronorm{\Sigstar - \Sighat}/\fronorm{\Sigstar}$ averaged over $10$ independent trials. For each query response, items are drawn i.i.d. from a standard multivariate normal distribution, similar to~\cite{massimino2021you}.

\paragraph{Pairwise comparison setup.} For pairwise comparisons, we use value of $\boundary = 10$ to denote the squared distance at which items become dissimilar, following our distance-based model for human perception (see Section~\ref{sec:paq:model}). For the $i$-th pairwise comparisons, we draw two items $\vx_{1}^{(i)}, \vx_{2}^{(i)}$ i.i.d. from a standard multivariate normal distribution. We record the pairwise comparison outcome $\epsilon_i \in \{-1, +1\}$ as $\epsilon_i = \text{sign}(\normsigstar{\vx_{1}^{(i)} - \vx_{2}^{(i)}}^2 - y)$. To estimate the metric from pairwise comparisons, we utilize a nuclear-norm regularized hinge loss and solve the following optimization problem:
\begin{align*}
    \Sighat_{\text{PC}} \in \argmin_{\mSigma \succeq \vzero} \frac{1}{N} \sum\limits_{i=1}^N \max\{0, y - \epsilon_i \normsig{\vx_{1}^{(i)} - \vx_{2}^{(i)}}^2 \} + \lambda_{\text{PC}}\nucnorm{\mSigma}.
\end{align*}

\paragraph{Triplet setup.} For the $i$-th triplet, we draw three items $\vx_{1}^{(i)}, \vx_{2}^{(i)}, \vx_{3}^{(i)}$ i.i.d. from a standard multivariate normal distribution and record the outcome $\epsilon_i \in \{-1, +1\}$ as $\epsilon_i = \text{sign}(\normsigstar{\vx_{1}^{(i)} - \vx_{2}^{(i)}}^2 - \normsigstar{\vx_{1}^{(i)} - \vx_{3}^{(i)}}^2)$. To estimate the metric from triplet responses, we follow~\cite{mason2017learning} and utilize a nuclear-norm regularized hinge loss and solve the following optimization problem:
\begin{align*}
    \Sighat_{\text{T}} \in \argmin_{\mSigma \succeq \vzero} \frac{1}{N} \sum\limits_{i=1}^N \max\left\{0, 1 - \epsilon_i\left( \normsig{\vx_{1}^{(i)} - \vx_{2}^{(i)}}^2 - \normsig{\vx_{1}^{(i)} - \vx_{3}^{(i)}}^2\right) \right\} + \lambda_{\text{T}}\nucnorm{\mSigma}.
\end{align*}

\paragraph{Ranking-$k$ query setup.} For the $i$-th ranking query with a reference item $\vx_0$ and $k$ items $\vx_1, \ldots, \vx_k$ to be ranked, we draw all items i.i.d. from a standard multivariate normal distribution. For each item $\vx_k$, we compute the squared distance $\normsigstar{\vx_0 - \vx_k}^2$. To determine the ranking of items, we sort the items based on this squared distance. To estimate the metric, we follow the approach of~\cite{canal2020tuplewise} and decompose the full ranking into its constituent triplets. A ranking consisting of $k$ items can equivalently be decomposed into $k(k-1)/2$ triplet responses. To estimate the metric, we decompose each ranking query and use the triplet estimator presented above with regularization parameter $\lambda_{\text{R}}$ to obtain estimate $\Sighat_{\text{R-}k}$.

\paragraph{\queryabbr setup.} For the $i$-th PAQ response, we draw the reference item $\vx_i$ and query vector $\va_i$ i.i.d. from the standard multivariate normal distribution. We then receive a scaling $\gamma_i^2$ satisfying $\gamma_i^2 = y / \quadSig{i}$, with $\boundary = 10$. To perform estimation, we leverage our method presented in Section~\ref{sec:estimator}. Our theoretical results indicate that the averaging parameter $m$ should be set to $1$ in the noiseless setting. Furthermore, the truncation threshold $\trunc$ is large relative to our responses $\gamma_i^2$, meaning no truncation is employed. As a result, we solve the nuclear-norm regularized trace regression problem
\begin{align*}
     \Sighat_{\text{PAQ}} \in \argmin_{\mSigma \succeq \vzero} \frac{1}{N}\sum\limits_{i=1}^N\left(\ip{\va_i\va_i^\top}{\mSigma} - \frac{y}{\gamma_i^2}\right)^2 + \lambda_{\text{PAQ}}\nucnorm{\mSigma}.
\end{align*}
\medskip

In all cases above, we solve all optimization problems with \texttt{cvxpy} and normalize the estimated metric $\Sighat_{\{\text{PC, T, R-$k$, PAQ}\}}$ to be unit Frobenius norm to ensure consistent scaling when compared against the true metric $\Sigstar$. We use a value of $0.05$ for all regularization parameters $\lambda_{\{\text{PC, T, R-$k$, PAQ}\}}$ and observe similar performance trends for other choices of regularization parameter.

\section{Scale equivariance}\label{app:scale-eq}
In this section, we verify that the scale-equivariance of our derived theoretical bounds~\eqref{eq:rate} and~\eqref{eq:rate-sqrt}. Specifically, we denote by $\Sigstar$ and $\Sighat$ the ground-truth and the estimated matrices corresponding to value $\boundary$. We denote by $\Sigstar_\const$ and $\Sighat_\const$ the ground-truth and estimated matrices corresponding to value $\constscale\boundary$ for any $\constscale > 0$. By definition, we have $\Sigstar_\const = \constscale\Sigstar$, and it can be verified that solving the optimization program~\eqref{eq:estimator} yields $\Sighat_\const = \constscale\Sighat$. Hence, one expects the error bound to scale as $\constscale$. To verify this linear scaling in $\constscale$, we confirm that the noise $\noise$ scales as $\constscale$. 

Under the ground-truth metric $\Sigstar$, if the user responds with an item that is a distance $\boundary + \noise$ away from the reference item, then that same item is a distance $\constscale(\boundary + \noise)$ away from the reference under the scaled setting. As a result, the noise scales as a result of the choice of $\boundary$. Therefore, the following values in the upper bound~\eqref{eq:rate} can be written as scaled versions of their corresponding ``ground-truth'' values. 

\begin{center}
\begin{tabular}{ l c c l c }
 Noise & $\noise = \constscale\  \noisestar$ & \text{    } & Noise median & $\noisemedian = \constscale\ \noisemedianstar$ \\ 
 Noise upper bound & $\noiseup = \constscale\ \noiseupstar$& \text{    } & Boundary upper bound & $\boundaryup = \constscale\ (\boundarystar + \noiseupstar)$ \\  
 Noise variance  & $\noisevar = \constscale^2\  \noisevarstar$ & \text{    } & Singular values & $\sigma_k = \constscale\ \sigma_k^{\star}, k = 1,\ldots,r$   
\end{tabular}
\end{center}
Substituting these scaled expressions into the upper bounds~\eqref{eq:rate} and~\eqref{eq:rate-sqrt}, we have
\begin{align*}
     \fronorm{\Sighat_\const - \Sigstar_\const} \leq \constscale\ C^\prime\ \frac{(\sigma_1^{\star})^2}{\sigma_r^{\star}} \frac{(\boundaryupstar)^{\nicefrac43} (\noisevarstar)^{\nicefrac13}}{(\noisemedianstar)^2}\ r^{\nicefrac{3}{2}}\left(\frac{d}{N} \right)^{\nicefrac{1}{3}}
\end{align*}
in the high-noise regime and
\begin{align*}
    \fronorm{\Sighat_\const - \Sigstar_\const} \leq \constscale\ C^\prime\ \frac{(\sigma_1^{\star})^2}{\sigma_r^{\star}} \left(\frac{\boundaryupstar}{\noisemedianstar}\right)^2r^{\nicefrac{3}{2}}\left(\frac{d}{N}\right)^{\nicefrac{1}{2}}
\end{align*}
in the low-noise regime. Note that the constant $\Const'$ is independent of $\constscale$. 

\section{Background and preliminary results}\label{app:prelim}
In this section, we provide an overview of the key tools that are utilized in our proofs.

\subsection{Inverted measurement sensing matrices result in estimation bias}\label{app:prelim:not_cond_mean_zero}
Recall from Equation~\eqref{eq:im_sensing_matrix} that the random sensing matrix $\mAinv$ takes the form
\begin{align*}
    \mAinv = \frac{\boundary + \noise}{\quadSig{}}\opa{}.
\end{align*}
Standard trace regression analysis assumes that for some sensing matrix $\mA$ and measurement noise $\noise$, $\E{\noise \mA} = \vzero$. Specifically, it is often typically assumed that $\noise$ is zero-mean conditioned on the sensing matrix $\mA$. The following lemma shows that for the inverted measurements, we have $\Expect[\noise\mAinv] \ne 0$, resulting in bias in estimation.
\begin{lemma}\label{lemma:not_conditional_mean_zero}
    Let $\mAinv$ be the random matrix defined in Eq.~\eqref{eq:im_sensing_matrix} and $\noise$ be the measurement noise. Then
    \begin{align*}
        \E{\noise \mAinv} \neq \vzero.
    \end{align*}
\end{lemma} 
The proof of Lemma~\ref{lemma:not_conditional_mean_zero} is provided in Appendix~\ref{app:prelim_proof:not_cond_mean_zero}. Hence, utilizing established low-rank matrix estimators for inverted measurements result in biased estimation. 

\subsection{Sub-exponential random variables}\label{app:prelim:sub_exp}
Our analysis utilizes properties of sub-exponential random variables, a class of random variables with heavier tails than the Gaussian distribution. 

\begin{lemma}[Moment bounds for sub-exponential random variables {\cite[Proposition 2.7.1(b)]{vershynin2018high}}]\label{lemma:subexp_moments}
    If $X$ is a sub-exponential random variable, then there exists some constant $c$ (only dependent on the distribution of the random variable $X$) such that for all integers $p \geq 1$,
    \begin{align*}
        \left(\Expect \abs*{X}^p\right)^{\nicefrac{1}{p}} \leq cp. 
    \end{align*}
\end{lemma}


\subsection{Bernstein's inequality} 
In our proofs, we use Bernstein's inequality to bound the sums of independent sub-exponential random variables. 

\begin{lemma}[Bernstein's inequality, adapted from {\cite[Theorem 2.10]{boucheron2013concentration}}]\label{lemma:bernsteins_inequality}
    Let $X_1, \ldots, X_n$ be independent real-valued random variables. Assume there exist positive numbers $\constbernone$ and $\constberntwo$ such that 
    \begin{align*}
        \E{X_i^2} \leq \constbernone \quad \text{ and } \quad \E{\abs*{X_i}^p} \leq \frac{p!}{2}\constbernone \constberntwo^{p-2} \text{ for all integers } p \ge 2,
    \end{align*}
    Then for all $t > 0$, 
    \begin{align*}
        \P{\abs*{\frac{1}{n}\sum\limits_{i=1}^n \left(X_i - \E{X_i}\right)} \geq \sqrt{\frac{2\constbernone t}{n}} + \frac{\constberntwo t}{n} } \leq 2\exp(-t).
    \end{align*}
\end{lemma}


\subsection{Moments of the ratios of quadratic forms}\label{app:prelim:bao_kan}
The quadratic term $\quadSig{}$ appears in the denominator of our sensing matrices, so we use the following result to quantify the moments of the ratios of quadratic forms.

\begin{lemma}\label{lemma:quad_form_moments}
There exists an absolute constant $\const > 0$ such that the following is true.
Let $\va\sample \normal(\vzero, \idmtx_\dimension)$, $\Sigstar\in \reals^{\dimension\times \dimension}$ be any PSD matrix with rank $\rank$, and $\mU \in \reals^{\dimension\times \dimension}$ be an arbitrary symmetric matrix.
\begin{enumerate}[label={(\alph*)}]
    \item \label{part:quad_form_inverse}
    Suppose that $r >8$. Then we have
    \begin{align*}
        \Expect\left(\frac{1}{\va^T \Sigstar \va}\right)^4 \le \frac{\const}{\singularvalmin^4 \rank^4}. 
    \end{align*}
    
    \item \label{part:quad_form_ratio}
    Suppose that $r > 2$. Then we have
    \begin{align*}
        \Expect\left(\frac{\va^\top \varmtx \va}{\va^\top \Sigstar \va}\right) \le \frac{\const}{\singularvalmin \rank}\nucnorm{\varmtx}.
    \end{align*}
\end{enumerate}
\end{lemma}

The proof of Lemma~\ref{lemma:quad_form_moments} is presented in Appendix~\ref{app:prelim_proof:quad_form}. 

\subsection{A fourth moment bound for \texorpdfstring{$\scalebar^2$}{squared scalings}}\label{app:prelim:fourth_moment}
Recall from Equation~\eqref{eq:avg_measurement} that the averaged measurement $\scalebar^2$ takes the form
\begin{align*}
    \scalebar_i^2 = \frac{1}{m}\sum_{j=1}^m \frac{\boundary + \noise_\idxitem^{(j)}}{\quadSig{\idxitem}}
    = \frac{\boundary + \noisebar_i}{\quadSig{i}}.
\end{align*}
Throughout our analysis, we utilize the fact that $\scalebar^2$ has a bounded fourth moment, as characterized in the following lemma.

\begin{lemma}\label{lemma:gamma_bar_fourth_moment}
    Assume $r > 8$. Then there exists a universal constant $\const > 0$, such that 
    \begin{align*}
        \Expect\left( \scalebar^2 \right)^4 \leq \const \left(\frac{\boundary + \noiseup}{\sigma_r r}\right)^4,
    \end{align*}
    where $\sigma_r$ is the smallest non-zero singular value of $\Sigstar$. 
\end{lemma}
The proof of Lemma~\ref{lemma:gamma_bar_fourth_moment} is presented in Appendix~\ref{app:prelim_proof:gamma_bar_fourth_moment}.
For notational simplicity of the proofs, we denote $\fourthmoment = c \left(\frac{\boundary + \noiseup}{\sigma_r r}\right)^4$.


\subsection{Proofs of preliminary lemmas}
In this section, we present proofs for preliminary lemmas from Appendices~\ref{app:prelim:not_cond_mean_zero},~\ref{app:prelim:bao_kan}, and~\ref{app:prelim:fourth_moment}.

\subsubsection{Proof of Lemma~\ref{lemma:not_conditional_mean_zero}}\label{app:prelim_proof:not_cond_mean_zero}
Using the independence of the noise $\noise$ and the sensing vector $\va$, and the assumption that $\noise$ is zero mean, we have 
\begin{align}
    \E{\noise \mAinv} &= \E{\frac{\noise(\boundary + \noise)}{\quadSig{}}\opa{}} \nonumber\\
    &= \E{\noise(\boundary + \noise)} \cdot\E{\frac{1}{\quadSig{}}\opa{}} \nonumber\\
    &= \noisevar\ \E{\frac{1}{\quadSig{}}\opa{}}.\label{eq:not_cond_mean_zero_expression}
\end{align}
The expectation in~\eqref{eq:not_cond_mean_zero_expression} is non-zero, because the random matrix $\frac{1}{\quadSig{}}\opa{}$ is symmetric positive definite almost surely. Therefore, we have $\E{\noise \mAinv} \neq \vzero$, as desired.

\subsubsection{Proof of Lemma~\ref{lemma:quad_form_moments}}\label{app:prelim_proof:quad_form}

Since $\Sigstar$ is symmetric positive semidefinite, it be decomposed as $\mQ\mSigma\mQ^\top$, where $\mQ$ is a square orthonormal matrix and $\mSigma$ is a diagonal matrix with non-negative entries. Multiplying $\va$ by any square orthonormal matrix does not change its distribution. Therefore, without loss of generality, we assume that $\Sigstar$ is diagonal with all non-negative diagonal entries. We first note that the moments of the ratios in both parts of Lemma~\ref{lemma:quad_form_moments} exist, because by~\cite[Proposition 1]{bao2013moments}, for non-negative integers $p$ and $q$, the quantity $\Expect \frac{\left(\quadmU{}\right)^p}{\left(\quadSig{}\right)^q}$ exists if $\frac{r}{2} > q$. Furthermore, we use the following expression from~\cite[Proposition 2]{bao2013moments}:
\begin{align}\label{eq:bao_kan_integral}
    \Expect \frac{\left(\quadmU{}\right)^p}{\left(\quadSig{}\right)^q}
    = \frac{1}{\Gamma(q)} \int\limits_{0}^\infty t^{q-1} \cdot \left| \mDelta_t \right| \cdot\Expect \left(\va^\top \mDelta_t \mU \mDelta_t \va\right)^p \dd t,
\end{align}
where $\mDelta_t = (\mId_d + 2t\Sigstar)^{-\nicefrac{1}{2}}$ and $\left|\mDelta_t\right|$ is the determinant of $\mDelta_t$. To characterize the determinant $\abs*{\mDelta_t}$, we note that $\mDelta_t$ is a diagonal matrix whose $\dimension$ diagonal entries are
\begin{align*}
    \frac{1}{(1 + 2t\sigma_1)^{\nicefrac{1}{2}}},\; \ldots,\; \frac{1}{(1 + 2t\sigma_r)^{\nicefrac{1}{2}}} , \; 1, \;\ldots, \; 1.
\end{align*}

Hence, the determinant is the product $\left|\mDelta_t\right| = \prod_{i=1}^r \frac{1}{(1 + 2t\sigma_i)^{\nicefrac12}}$. Furthermore, this product can be bounded as:
\begin{align}
    \left|\mDelta_t\right| \leq \frac{1}{(1 + 2t\sigma_r)^{\nicefrac{r}{2}}} \label{eq:bao_kan_det_bound}.
\end{align}
We  now prove parts (a) and (b) separately.

\paragraph{Part (a).} 
Using the integral expression~\eqref{eq:bao_kan_integral} with $p = 0$ and $q = 4$, and the upper bound~\eqref{eq:bao_kan_det_bound} on the determinant, we have
\begin{align*}
    \Expect\left(\frac{1}{\quadSig{}}\right)^4 &= \frac{1}{\Gamma(4)} \int\limits_{0}^\infty t^{3} \cdot \abs*{\mDelta_t} \dd t \\
    &\leq \frac{1}{\Gamma(4)} \int\limits_{0}^\infty t^{3} \frac{1}{(1 + 2t\sigma_r)^{\nicefrac{r}{2}}} \dd t.
\end{align*}
Denoting $s \defn 1 + 2t\sigma_r$, we have
\begin{align*}
    \Expect \left(\frac{1}{\quadSig{}}\right)^4 &\leq \frac{1}{2 \Gamma(4) \sigma_r} \int\limits_{1}^\infty \left(\frac{s-1}{2\sigma_r}\right)^3 \frac{1}{s^{\nicefrac{r}{2}}} \dd s \\
    &\lesssim \frac{1}{\sigma_r^4 } \int\limits_{1}^\infty \frac{(s-1)^3}{s^{\nicefrac{r}{2}}} \dd s\\
    &= \frac{1}{\sigma_r^4} \int\limits_{1}^\infty \left(\frac{s^3}{s^{\nicefrac{r}{2}}} - 3\frac{s^2}{s^{\nicefrac{r}{2}}} + 3\frac{s}{s^{\nicefrac{r}{2}}} - \frac{1}{s^{\nicefrac{r}{2}}}\right) \dd s\\
    &= \frac{1}{\sigma_r^4} \left(\frac{2}{r-8} - \frac{6}{r-6} + \frac{6}{r-4} - \frac{2}{r-2} \right)\\
    & \le \frac{\const}{\singularvalmin^4 \rank^4},
\end{align*}
as desired.

\paragraph{Part (b).}
Using the integral expression~\eqref{eq:bao_kan_integral} $p = q = 1$ and the upper bound~\eqref{eq:bao_kan_det_bound} on the determinant, we have
\begin{align}
    \Expect \left(\frac{\quadmU{}}{\quadSig{}}\right) &= \frac{1}{\Gamma(1)} \int\limits_{0}^\infty \abs*{\mDelta_t} \cdot \E{\va^\top \mDelta_t \mU \mDelta_t \va} \dd t \nonumber\\ 
    &\leq \frac{1}{\Gamma(1)} \int\limits_{0}^\infty \frac{1}{(1 + 2t\sigma_r)^{\nicefrac{r}{2}}} \E{\va^\top \mDelta_t \mU \mDelta_t \va} \dd t.\label{eq:ratio_part_two_expression}
\end{align}
We now bound the expectation term in~\eqref{eq:ratio_part_two_expression}. Note that for $\va \sim \normal(\vzero, \mId_d)$, we have $\E{\va^\top \mB \va} = \tr{\mB}$ for any symmetric matrix $\mB$. Therefore, we have
\begin{align}
    \E{\va^\top \mDelta_t \mU \mDelta_t \va} &= \tr{\mDelta_t \mU \mDelta_t}\nonumber\\
    &\overset{\stepone}{\leq} \nucnorm{\mDelta_t \mU \mDelta_t} \nonumber\\
    &\overset{\steptwo}{\leq} \nucnorm{\mU},\label{eq:ratio_part_two_expt_bound}
\end{align}
where \stepone the fact that $\tr{\mB} \leq \nucnorm{\mB}$ for any symmetric matrix $\mB$. Furthermore, \steptwo follows from H\"{o}lder's inequality for Schatten-$p$ norms, where we have that $\nucnorm{\mDelta_t \mU \mDelta_t} \leq \opnorm{\mDelta_t}^2 \cdot \nucnorm{\mU}$. Because $\mDelta_t$ is diagonal and the entries of $\mDelta_t$ are bounded between $0$ and $1$, we bound the operator norm as $\opnorm{\mDelta_t} \leq 1$. Substituting~\eqref{eq:ratio_part_two_expt_bound} to~\eqref{eq:ratio_part_two_expression}, we obtain
\begin{align*}
    \Expect\left( \frac{\quadmU{}}{\quadSig{}}\right) & \leq \nucnorm{\mU}\cdot \int\limits_{0}^\infty \frac{1}{(1 + 2t\sigma_r)^{\nicefrac{r}{2}}} \dd t\\
    & \lessorder  \frac{1}{\sigma_r r}\cdot \nucnorm{\mU},
\end{align*}
as desired.

\subsubsection{Proof of Lemma~\ref{lemma:gamma_bar_fourth_moment}}\label{app:prelim_proof:gamma_bar_fourth_moment}
By the assumption that the noise is upper bounded by $\noiseup$, we have $\boundary + \noisebar \leq \boundary + \noiseup$. Therefore, we have
\begin{align*}
    \Expect \left(\scalebar^2\right)^4 &= \Expect\left( \frac{y+\noisebar}{\quadSig{}} \right)^4\\
    &\leq (\boundary+\noiseup)^4\cdot \Expect\left( \frac{1}{\quadSig{}} \right)^4\\
    &
    \stackrel{\stepone}{\lesssim} \left(\frac{1}{\sigma_r r}\right)^4,
\end{align*}
where step~\stepone applies part~\ref{part:quad_form_inverse} of \Cref{lemma:quad_form_moments}.

\end{document}